\documentclass[journal,transmag]{IEEEtran}
\usepackage{mathtools}
\usepackage[pdftex]{graphicx} 
\usepackage{microtype}
\usepackage{amsmath,amssymb}
\usepackage[table,pdftex]{xcolor}
\usepackage[caption=false,font=footnotesize]{subfig} 
\usepackage[british]{babel}
\usepackage{array}
\usepackage{multirow}
\usepackage{fixltx2e} 
\usepackage{todonotes}
\usepackage{algorithm}
\usepackage{algpseudocode}
\usepackage{hyperref}

\usepackage{booktabs}
\usepackage{threeparttable}
\usepackage{footnote}
\usepackage{bbm}

\title{Automatic 3D bi-ventricular segmentation of cardiac images by a shape-refined multi-task deep learning approach}

\author{Jinming Duan, Ghalib Bello, Jo Schlemper, Wenjia Bai, Timothy J W Dawes, Carlo Biffi, \\Antonio de Marvao, Georgia Doumou, Declan P O'Regan, Daniel Rueckert,~\IEEEmembership{Fellow,~IEEE}
\thanks{JD, JS, WB, CB and DR are with the Biomedical Image Analysis Group, Imperial College London, London, UK. JD, GB, TJWD, A de M, GD and DPO'R are with MRC London Institute of Medical Sciences, Imperial College London, London, UK. TJWD is with the National Heart and Lung Institute, Imperial College London, London, UK.}}
\date{}

\begin{document}
\maketitle

\begin{abstract}

\noindent Deep learning approaches have achieved state-of-the-art performance in cardiac magnetic resonance (CMR) image segmentation. However, most approaches have focused on learning image intensity features for segmentation, whereas the incorporation of anatomical shape priors has received less attention. In this paper, we combine a multi-task deep learning approach with atlas propagation to develop a shape-refined bi-ventricular segmentation pipeline for short-axis CMR volumetric images. The pipeline first employs a fully convolutional network (FCN) that learns segmentation and landmark localisation tasks simultaneously. The architecture of the proposed FCN uses a 2.5D representation, thus combining the computational advantage of 2D FCNs networks and the capability of addressing 3D spatial consistency without compromising segmentation accuracy. Moreover, a refinement step is designed to explicitly impose shape prior knowledge and improve segmentation quality. This step is effective for overcoming image artefacts (e.g. due to different breath-hold positions and large slice thickness), which preclude the creation of anatomically meaningful 3D cardiac shapes. The pipeline is fully automated, due to network's ability to infer landmarks, which are then used downstream in the pipeline to initialise atlas propagation. We validate the pipeline on 1831 healthy subjects and 649 subjects with pulmonary hypertension. Extensive numerical experiments on the two datasets demonstrate that our proposed method is robust and capable of producing accurate, high-resolution and anatomically smooth bi-ventricular 3D models, despite the presence of artefacts in input CMR volumes.

\end{abstract}

 

\section{Introduction}\label{sec:introduction}  

\IEEEPARstart{C}{ardiac} magnetic resonance (CMR) imaging is the gold standard for assessing cardiac chamber volume and mass for a wide range of cardiovascular diseases \cite{ripley2016cardiovascular}. For decades, clinicians have been relying on manual segmentation approaches to derive quantitative measures such as left ventricle (LV) volume, mass and ejection fraction. However, manual expert segmentation of CMR images is tedious, time-consuming and prone to subjective errors. It becomes impractical when dealing with large-scale datasets. As such, there is a demand for automatic techniques for CMR image analysis that can handle the scale and variability associated with large imaging studies \cite{medrano2015challenges,rueckert2016learning}. Recently, automatic segmentation based on deep neural networks has achieved state-of-the-art performance in the CMR domain \cite{winther2017nu,patravali20172d,baumgartner2017exploration,isensee2017automatic,zheng20183d,nasr2018left,khened2018fully,oktay2018anatomically,tran2016fully,bai2017human,ngo2017combining,avendi2016combined,duan2018deep, schlemper2018cardiac}. For example, in the Automatic Cardiac Diagnosis Challenge (ACDC) \cite{bernard2018deep} the 8 highest-ranked segmentation methods were all neural network-based methods. 

Theoretically, 3D neural network-based segmentation methods may be designed with arbitrarily deep architectures. In practice however, the size of cardiac images, especially that of high-resolution volumetric images \cite{oktay2018anatomically}, often presents a computational bottleneck at the training stage. To deal with this, shallow 3D network architectures \cite{oktay2018anatomically} or fewer feature/activation maps \cite{patravali20172d} are typically considered. Also, to reduce the computational burden, most methods extract the region of interest (ROI) containing the whole heart as a first step to reduce the volume size \cite{zheng20183d,nasr2018left,khened2018fully,oktay2018anatomically,ngo2017combining,avendi2016combined, schlemper2018cardiac}, or train a 2D network to separately segment each short-axis slice in the volume \cite{tran2016fully,bai2017human,ngo2017combining,avendi2016combined,duan2018deep}. However, there are fundamental problems associated with each of these workarounds. For example, the use of shallow 3D network architectures or fewer feature maps is known to compromise segmentation accuracy. The ROI extraction approach is carried out using ROI detection algorithms, whose robustness remains questionable \cite{zheng20183d}. In addition, as no 3D context is taken into account, 2D network-based methods suffer from lack of 3D spatial consistency between the segmented slices (leading to lack of smoothness in the long-axis direction), and may result in a false positive prediction at an image slice containing non-ventricular tissues that are similar to target ventricles \cite{zheng20183d}.

\begin{figure}[h!] 
\centering
\includegraphics[width=0.48\textwidth]{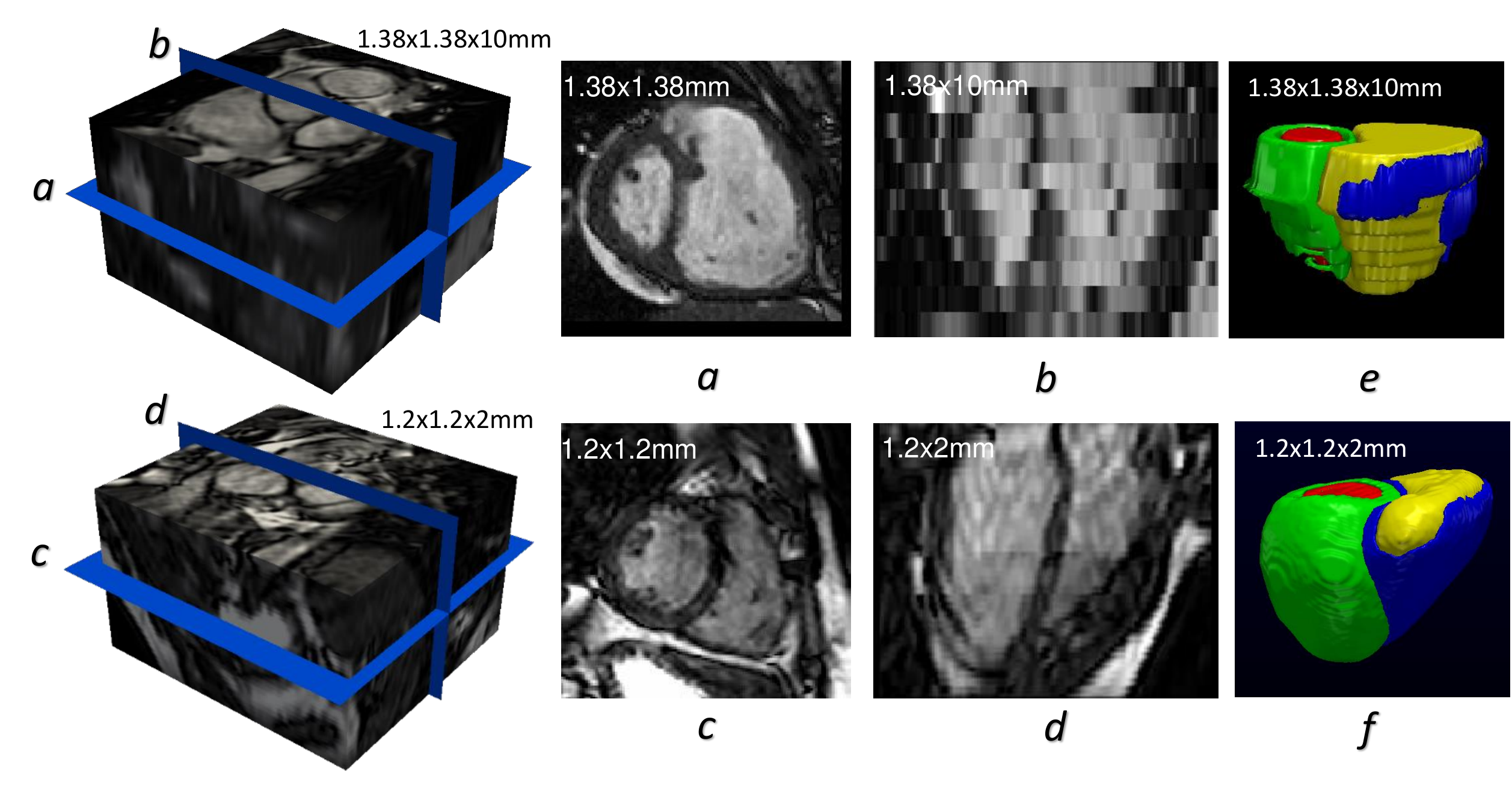}\\
\vspace{-7pt}
\caption{Illustrating the differences between a low-resolution CMR volume (top row) and a high-resolution CMR volume (bottom row). The images in the short-axis view are shown in $a$ and $c$, while those in the long-axis view are in $b$ and $d$.  The corresponding segmentations are given in $e$ and $f$.}
\label{fig:LRHR}
\vspace{-5pt}
\end{figure}

Due to the limitations of standard clinical acquisition protocols, raw volumetric CMR images acquired from standard scans often contain several artefacts \cite{petersen2015uk}, including inter-slice shift (i.e. respiratory motion), large slice thickness, and lack of slice coverage. Most deep learning methods do not routinely account for imaging artefacts \cite{winther2017nu,patravali20172d,baumgartner2017exploration,isensee2017automatic,zheng20183d,nasr2018left,khened2018fully,tran2016fully,bai2017human,ngo2017combining,duan2018deep}. As such, these artefacts are inevitably propagated onto the resulting segmentations. An example is given in Fig~\ref{fig:LRHR} $e$. The figure shows the segmentation of a 3D volume (whose short- and long-axis views are shown in Fig~\ref{fig:LRHR} $a$ and $b$) using a state-of-the-art CNN approach \cite{bai2017human}. As can be seen, the segmentation Fig~\ref{fig:LRHR} $e$ inherits the misalignment and staircase artefacts present in the original volumetric image due to cardiac motion and large slice thickness. Further, holes exist at the apical region of the 3D model due to incomplete slice coverage of the whole heart. Different approaches have been proposed to tackle each artefact accordingly before building a smooth model. For example, misalignment was corrected using quadratic polynomials \cite{avendi2016combined} or rigid registration \cite{tarroni2018learning}; Large slice thickness can be addressed by super-resolution techniques \cite{oktay2016multi}. However, few studies have addressed different artefacts directly from an image segmentation perspective. To date, we are aware of only one deep learning segmentation method \cite{oktay2018anatomically} that takes into account different cardiac artefacts, but the method was tested on only simulated images of the LV, whose anatomy is less complex than the bi-ventricular anatomy. It is thereby still an open problem as to how to build an artefact-free and smooth bi-ventricular segmentation model from real artefact-corrupted CMR volumes with novel image segmentation methods. 

For clinical applications, segmentation algorithms need to maintain accuracy across diverse patient populations with varying disease phenotypes. In the existing literature, however, most methods \cite{patravali20172d,baumgartner2017exploration,nasr2018left,oktay2018anatomically,tran2016fully,bai2017human,ngo2017combining,avendi2016combined} have been developed and validated over normal (healthy) hearts or mildly abnormal hearts. Few studies have focused on hearts with very significant pathology with altered geometry and motion compared to healthy hearts. 
In addition,  most methods \cite{winther2017nu,patravali20172d,baumgartner2017exploration,isensee2017automatic,nasr2018left,khened2018fully,tran2016fully,ngo2017combining,avendi2016combined} tend to use small image datasets. For example, four representative MICCAI challenges, namely the 2009 automatic LV segmentation challenge\footnote{\href{http://www.cardiacatlas.org/challenges/lv-segmentation-challenge/}{http://www.cardiacatlas.org/challenges/lv-segmentation-challenge/}} (also known as Sunnybrook cardiac data), the 2011 LV segmentation challenge\footnote{\href{http://www.cardiacatlas.org/studies/sunnybrook-cardiac-data/}{http://www.cardiacatlas.org/studies/sunnybrook-cardiac-data/}} (organized as part of the STACOM workshop), the 2015 RV segmentation challenge \cite{petitjean2011review} and the 2017 ACDC, were tested on only 30, 100, 48 and 100 CMR datasets respectively. Given the small size of the datasets used for training and testing, whether the reported results can be generalised to larger cohorts remains questionable.

In this paper, we propose a segmentation pipeline to address the aforementioned limitations of current approaches. Specifically, we make the following contributions: 
\begin{itemize}
\item We propose a multi-task deep learning network that simultaneously predicts segmentation labels and anatomical landmarks in CMR volumes. The network takes input volumetric images as multi-channel vector images (2.5D representation), requires no ROI extraction, and contains up to 15 convolutional layers. As such, the network has the computational advantage of 2D networks and is able to address 3D issues without compromising accuracy and spatial consistency. To our knowledge, this is the first work applying deep learning to CMR landmark localisation in a 3D context.
\item We introduce anatomical shape prior knowledge to the network segmentation, which is a refinement step that is carried out using atlas propagation with a cohort of high-resolution atlases. As such, the pipeline is able to produce an accurate, smooth and clinically meaningful bi-ventricular segmentation model, despite the existing artefacts in the input volume. Moreover, due to the use of landmarks detected by the network, the proposed pipeline is entirely automatic.
\item We demonstrate that the proposed pipeline can be readily generalised to segmenting volumetric CMR images from subjects with pulmonary hypertension (a cardiovascular disease). We thoroughly assess the effectiveness and robustness of the proposed pipeline using a large-scale dataset, comprising 2480 short-axis CMR volumetric images for training and testing. To our knowledge, this is one of the first CMR segmentation studies utilising a volumetric dataset of this size, and the technique introduced herein is the first automatic approach capable of producing a full high-resolution bi-ventricular model in 3D.
\end{itemize}

\section{Method}

\vspace{-10pt}
\subsection{Overview}
\label{overview}
The proposed automatic segmentation pipeline handles two types of CMR volumetric inputs: low-resolution (LR) and high-resolution (HR) volumes. Fig~\ref{fig:LRHR} illustrates the differences between them. The LR volume has a large slice thickness (10 mm), giving rise to a staircase effect in the long-axis\footnote{In a standard CMR acquisition, short-axis and long-axis images are acquired separately, both of which have high in-plane resolution. However, in this paper, only CMR-acquired short-axis images are used, and a long-axis image denotes a vertical slice/cross-section of a stack of these short-axis images. Large thickness between short-axis images would result in a poor resolution in the long-axis image. An example is given in Fig~\ref{fig:LRHR}.} view (Fig~\ref{fig:LRHR} $b$). Moreover, since the slices in Fig~\ref{fig:LRHR} $b$ were acquired from multiple breath-holds, inconsistency of each breath-hold results in an inter-slice shift artefact. In contrast, {the cross plane resolution }of the HR volume is 2 mm, making its long-axis image Fig~\ref{fig:LRHR} $d$ relatively smooth. In addition, HR imaging requires only one single 20-25 second breath-hold and therefore it introduces no inter-slice shift artefact. However, HR imaging may not be feasible for pathological subjects who are unable to hold their breath for 20-25s during each scan. Since HR imaging acquisition generates artefact-free cardiac volumes \cite{de2014population}, it enables an accurate delineation of ventricular morphology, as shown in Fig~\ref{fig:LRHR} $f$. In comparison, Fig~\ref{fig:LRHR} $e$ shows that the segmentation of an LR volume contains different cardiac artefacts \cite{petersen2015uk} (e.g. inter-slice shift, large slice thickness, and lack of slice coverage). Note that the in-plane resolution of both HR and LR volumes is about $1.3 \times 1.3 $ mm, so their corresponding short-axis views Fig~\ref{fig:LRHR} $a$ and $c$ are of relatively high quality.

\begin{figure}[h!] 
\centering
\includegraphics[width=0.49\textwidth]{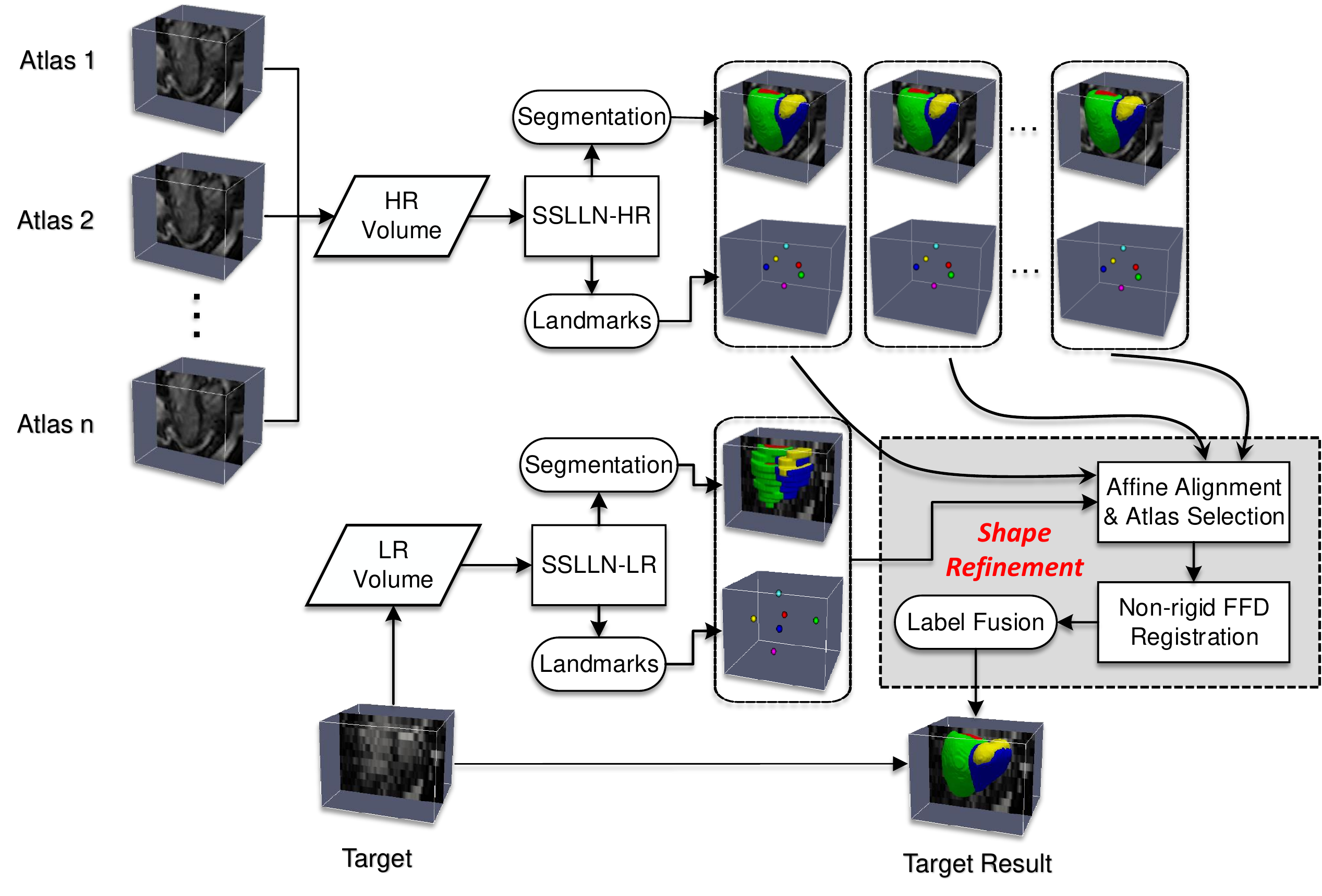}\\
\vspace{-7pt}
\caption{Pipeline for automatic bi-ventricular segmentation of low- and high-resolution volumetric images. The pipeline includes segmentation, landmark localisation and atlas propagation. It is capable of producing accurate, high-resolution and anatomically smooth bi-ventricular models, despite existing artefacts in input CMR volumes.} 
\label{fig:flowchart}
\vspace{-5pt}
\end{figure}

The proposed pipeline has three main components: segmentation, landmark localisation and atlas propagation. We term the proposed network used in the pipeline as the Simultaneous Segmentation and Landmark Localisation Network (SSLLN). Further, the related terms SSLLN-HR and SSLLN-LR will be used to refer to versions of SSLLN trained with HR and LR volumetric data, respectively. In Fig~\ref{fig:flowchart}, we illustrate the pipeline schematically. For an HR volume input, the trained SSLLN-HR is deployed to predict its segmentation labels as well as landmark locations. Since the HR volume input is artefact-free, the resulting segmentation is an accurate and smooth bi-ventricular 3D model. Afterwards, the HR volume and its corresponding SSLLN-HR outputs (landmarks and segmentation) are used as part of an HR atlas. For an LR volume input, the pipeline consists of two steps: First, the trained SSLLN-LR predicts an initial segmentation of the LR volume. In order to guarantee an artefact-free smooth segmentation output, a further refinement is carried out (second step). In this step, multiple selected HR atlases derived from SSLLN-HR are propagated onto the initial LR segmentation to form a smooth segmentation. This step explicitly fits anatomical shapes and is fully automatic due to the use of landmarks predicted from SSLLN-HR and -LR. We detail each of the two steps in the next two subsections. 

\subsection{Learning segmentation labels and landmark locations }
\begin{figure}[h!] 
\vspace{-5pt}
\centering
\includegraphics[width=0.45\textwidth]{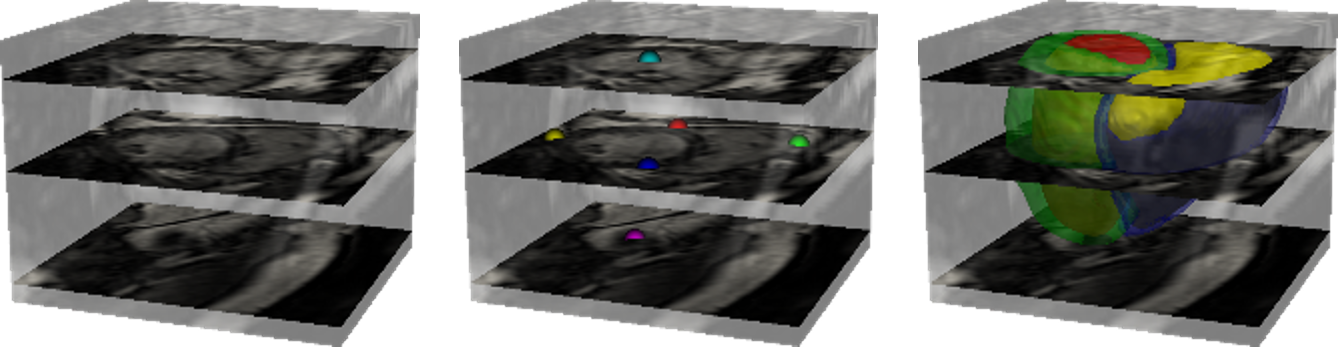}\\
\vspace{-7pt}
\caption{An exemplar raw volumetric CMR image, its ground-truth landmarks and segmentation labels, which are utilised as inputs to train the network in Fig~\ref{fig:network}. On the left, three short-axis slices in the volume are highlighted, corresponding to basal, mid-ventricular, and apical locations (from top to bottom) of the heart. In the middle, six landmarks are shown, coloured according to the following cardiac regions: the left ventricular lateral wall mid-point (yellow), two right ventricular insert points (red and blue), right ventricular lateral wall turning point (green), apex (pink) and centre of the mitral valve (cyan). Together, they reflect the size, pose and shape of the heart. On the right, a full anatomical bi-ventricular heart model is shown, coloured according to the left ventricular cavity (red), left ventricular wall (green), right ventricular cavity (yellow) and right ventricular wall (blue). }
\label{fig:exampleFig}
\end{figure}

We treat the problem of predicting segmentation labels and landmark locations as a multi-class classification problem. First, let us formulate the learning problem as follows: we denote the input volumetric training dataset by $S=\{(U_i, R_i, L_i), i=1,...,N_t\}$, where $U_i=\{u^i_j,j=1,...,|U_i|\}$ is the raw input CMR volume (Fig~\ref{fig:exampleFig} left), $R_i=\{r^i_j,j=1,...,|R_i|\}$, $r^i_j \in \{1,...,N_r\}$ denotes the ground-truth segmentation labels for volume $U_i$ ($N_r=5$ representing 4 tissue types and a background as shown in Fig~\ref{fig:exampleFig} right), $L_i=\{l^i_j,j=1,...,|L_i|\}$, $l^i_j \in \{1,...,N_l\}$ stands for the ground-truth landmark labels for $U_i$ ($N_l=7$ representing 6 landmarks and a background as shown in Fig~\ref{fig:exampleFig} middle), and $N_t$ is the number of samples in the training data.

Note that $|U_i|=|R_i|=|L_i|$ is the total number of voxels in a CMR volume. We then define all network layer parameters as $\textbf{W}$. In a supervised setting, we propose to solve the following minimisation problem via the standard (back-propagation) stochastic gradient descent
\begin{equation} \label{eq:SDLloss}
{{\bf{W}}^*} = \mathop {{\rm{argmin}}}\limits_{\bf{W}} ({L_D}({\bf{W}}) + \alpha {L_L}({\bf{W}}) + \beta \|{\bf{W}}\|_F^2),
\end{equation}
where $\alpha$ and $\beta$ are weight coefficients balancing the three terms. $L_D(\textbf{W})$ is the segmentation loss that evaluates spatial overlap with ground-truth labels. $L_L(\textbf{W})$ is the landmark associated loss for predicting landmark locations. $\|\textbf{W}\|_F^2$, known as the weight decay term, represents the Frobenius norm on the weights $\textbf{W}$. This term is used to prevent over-fitting in the network. The training problem is to estimate the parameters $\textbf{W}$ associated with all the convolutional layers and by minimising (\ref{eq:SDLloss}) the network is able to simultaneously predict segmentation labels and landmark locations. The definition of $L_D(\textbf{W})$ above is first given as follows 
\begin{equation} \label{eq:Dloss}
{L_D}({\bf{W}}) =  - \sum\limits_i {\frac{{2\sum\limits_k {\sum\limits_j {{\mathbbm{1}_{\left\{ {r_j^i = k} \right\}}} \cdot P(r_j^i = k|{U_i},{\bf{W}})} } }}{{\sum\limits_k {\sum\limits_j {\left( {\mathbbm{1}_{\left\{ {r_j^i = k} \right\}}^2 + {P^2}(r_j^i = k|{U_i},{\bf{W}})} + \epsilon \right) } } }}},
\end{equation} 
where $\mathbbm{1}_{\{\cdot\}}$ is an indicator function. $\epsilon$ is a small positive value used to avoid dividing by zero. $i$, $k$ and $j$ respectively denote the training sample index, the segmentation label index and the voxel index. $P(r^i_j=k|U_i,\textbf{W})$ corresponds to the softmax probability estimated by the network for a specific voxel $j$ (subject to the restriction $r_j^i=k$), given the training volume $U_i$ and network weights $\textbf{W}$. Note that (\ref{eq:Dloss}) is known as the differentiable Dice loss \cite{milletari2016v}, in which the summations are carried out over all voxels, labels and training samples.

For landmark localisation in a CMR volume, the primary challenge is the extreme imbalance between the proportion of voxels belonging to landmark regions and the proportion belonging to non-landmark regions (the 6 landmarks are represented by 6 voxels, while all the remaining voxels (numbering in the millions) represent background). To solve this highly imbalanced classification problem, we propose the class-balanced weighted categorical cross-entropy loss
\begin{equation} \label{eq:Lloss}
{L_L}({\bf{W}}) =  - \sum\limits_{i} {\sum\limits_{k} {\left( {w_k^i\sum\limits_{j \in {Y_k^i}} {{\rm{log}}P(l_j^i = k|{U_i},{\bf{W}})} } \right)}}.
\end{equation} 
Here $k$ denotes the landmark label index, ranging from 1 to 7. $Y_k^i$ represents the voxels in training sample $i$ that belong to the region for which the value of landmark label index is $k$. To automatically balance landmark and non-landmark classes, we use a weight $w_k^i$ for (\ref{eq:Lloss}), where $w_k^i = 1 - {{\left| {Y_k^i} \right|} \mathord{\left/{\vphantom {{\left| {Y_k^i} \right|} {\left| {{Y_i}} \right|}}} \right. \kern-\nulldelimiterspace} {\left| {{Y_i}} \right|}}$, $k=1,..,7$. Here $|Y_k^i|$ denotes the number of voxels in $Y_k^i$, while $|Y_i|$ represents the total number of voxels in training sample $i$. Let us explain how the weighting process works intuitively. For the voxel falling in any one of the 6 landmark locations, $\left| {Y_k^i} \right|$ is 1 and ${{\left| {Y_k^i} \right|} \mathord{\left/{\vphantom {{\left| {Y_k^i} \right|} {\left| {{Y_i}} \right|}}} \right. \kern-\nulldelimiterspace} {\left| {{Y_i}} \right|}}$ is close to zero. Therefore, $1 - {{\left| {Y_k^i} \right|} \mathord{\left/{\vphantom {{\left| {Y_k^i} \right|} {\left| {{Y_i}} \right|}}} \right. \kern-\nulldelimiterspace} {\left| {{Y_i}} \right|}}$ is close to 1. On the other hand, ${\sum\limits{{\rm{log}}P(l_j^i = k|{U_i},{\bf{W}})} }$ in (\ref{eq:Lloss}) is very small as only one voxel contributes to this term. Therefore, the product ${w_k^i\sum\limits{{\rm{log}}P(l_j^i = k|{U_i},{\bf{W}})} }$ ends up being a small value. In contrast, for a voxel falling in background area, $1 - {{\left| {Y_k^i} \right|} \mathord{\left/{\vphantom {{\left| {Y_k^i} \right|} {\left| {{Y_i}} \right|}}} \right. \kern-\nulldelimiterspace} {\left| {{Y_i}} \right|}}$ is a very small value close to zero. ${\sum\limits{{\rm{log}}P(l_j^i = k|{U_i},{\bf{W}})} }$ is however very large as almost all voxels (excluding the 6 landmark voxels) contribute to this term. Therefore, the product ${w_k^i\sum\limits{{\rm{log}}P(l_j^i = k|{U_i},{\bf{W}})} }$ becomes a small value. As such. the losses resulting from the landmark and non-landmark voxels are well balanced, which is crucial for successfully detecting merely 6 landmarks from a volume containing millions of voxels.
\begin{figure}[h!] 
\vspace{-10pt}
\centering
\includegraphics[width=0.49\textwidth]{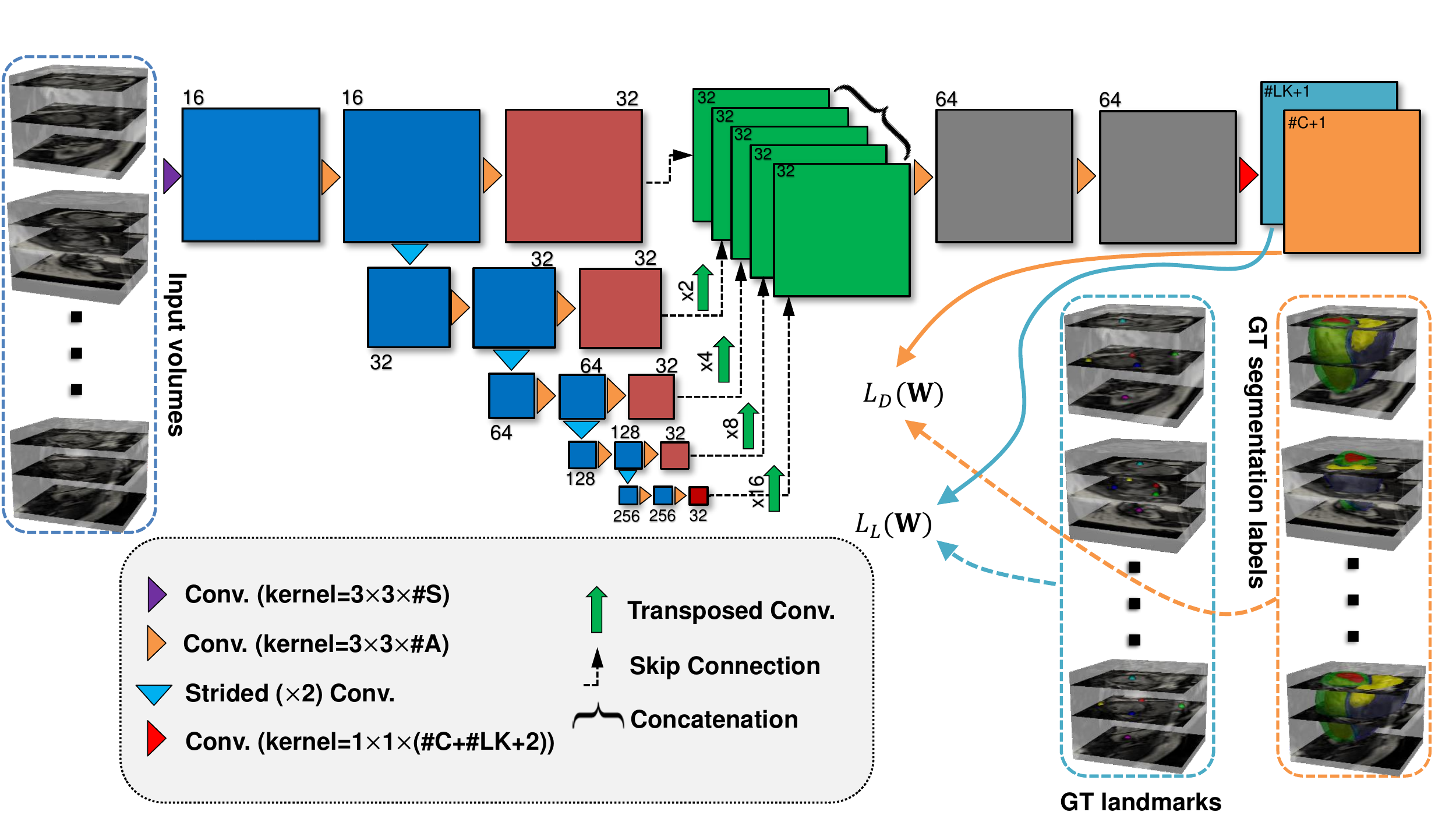}
\vspace{-8pt}
\caption{The architecture of the proposed SSLLN with 15 convolutional layers. The network takes different CMR volumes as input, applies a branch of convolutions, learns image features from fine to coarse levels, concatenates multi-scale features and finally predicts the probability maps of segmentation and landmarks simultaneously. These probability maps, together with the ground-truth segmentation labels and landmark locations, are then utilised in the loss function in (\ref{eq:SDLloss}) which is minimised via the stochastic gradient descent. Here $\#$S, $\#$A, $\#$C, $\#$LK and GT represent the number of volume slices, the number of activation maps, the number of anatomies, the number of landmarks, and ground truth, respectively.}
\label{fig:network}
\end{figure}

In Fig~\ref{fig:network}, we show the architecture of SSLLN. There are two major differences between our network architecture and existing 2D or 3D ones, which we highlight as novel contributions of this work. First, 2D networks \cite{winther2017nu,patravali20172d,baumgartner2017exploration,isensee2017automatic,nasr2018left,khened2018fully,tran2016fully,bai2017human,ngo2017combining,avendi2016combined,duan2018deep} are often trained using 2D short-axis slices separately. Therefore, there is no 3D spatial consistency between the resulting segmented slices. 3D networks \cite{patravali20172d,baumgartner2017exploration,isensee2017automatic,oktay2018anatomically,milletari2016v} often rely on 3D convolutions, which in practice leads to 5D tensors (e.g. batch size $\times$ [3D volume size] $\times$ classification categories) during forward and backward propagations and requires far more GPU memory than their 2D counterparts. Workarounds such as subsampling \cite{kamnitsas2017efficient} or use of small batch size and fewer convolutional layers \cite{patravali20172d,isensee2017automatic,oktay2018anatomically} are often considered when training 3D networks, but these either complicate the training process or cause loss of information and accuracy. Unlike 2D networks, our network treats each input CMR volume as a multi-channel vector image, known as `2.5D' representation. In this sense, 3D volumes rather than 2D short-axis slices are used to train our network. As such, our network accounts for the spatial consistency between slices. Retaining the 3D spatial relationship is crucial for landmark localisation as landmarks encode spatial information. Unlike 3D networks, our network only involves 4D tensors (excluding the last layer). After the input volume passes through the first convolutional layer, the subsequent convolutional operations (excluding the last layer) in our network function exactly the same as those in 2D methods. Hence, the proposed network has the computational advantage of 2D networks, and also handles the input explicitly as a 3D volume (rather than a series of 2D slices), thus retaining accuracy and spatial consistency. This will be demonstrated later in Section \ref{HRExp}. We also note that other network architecture, such as the multi-view CNN \cite{mortazi2017cardiacnet} that parses 3D data into different 2D components, may also suit our applications. Second, our network predicts segmentation labels and landmark locations simultaneously as we integrate the two problems into a unified image classification problem for which we tailored a novel loss function (\ref{eq:SDLloss}). We are not aware of any previous approach that detects cardiac landmarks using a deep learning-based classification method. This is also the first work that focuses on segmentation and landmark localisation simultaneously.

After the network is trained, given an unseen CMR volume $f : \Omega \to {\mathbb{R}^{\#S}}$ ($\#S$ is the number of short-axis slices in the volume) defined on the domain $\Omega \subset \mathbb{R}^2$, we deploy the network on it and obtain the probability maps of segmentation ($P_S$) and the probability maps of landmarks ($P_L$) from the last convolutional layer. The binary segmentation and landmark labels are the indices of the maximum values of their probability maps along the channel direction, i.e. $S = \arg {\max _{k = 1,...,{N_r}}}{P_S}$ and ${\cal L} = \arg {\max _{k = 1,...,{N_l}}}{P_L}$.
\subsection{Introducing anatomical shape prior knowledge}
\label{shapeconstraint}
\begin{figure*}[h!] 
\vspace{-5pt}
\centering
\includegraphics[width=0.99\textwidth]{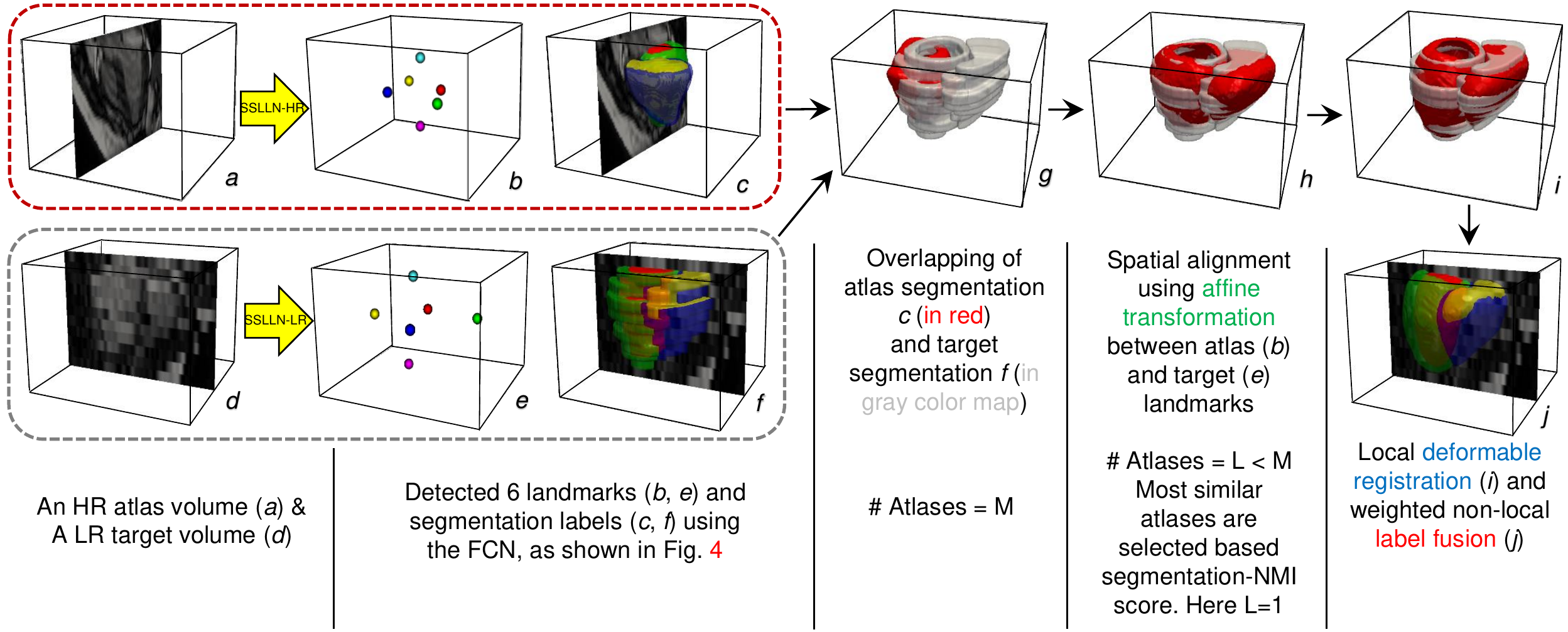}
\vspace{-10pt}
\caption{A block diagram illustrating how to explicitly introduce an anatomical shape refinement to the SSLLN-LR segmentation. As is evident in $j$, such a shape refinement enables an accurate, smooth and clinically meaningful bi-ventricular segmentation model, despite the artefacts in the LR input volume $d$. The framework is fully automated due to the use of the landmarks detected from SSLLN-HR and -LR.}
\label{fig:regisration}
\end{figure*}

Due to limitations of cardiac MR imaging, low-resolution (LR) volumetric training datasets often contain artefacts, such as inter-slice shift, large slice thickness, lack of slice coverage, etc. Inevitably, the deployment of SSLLN-LR trained from such a dataset causes the propagation of these artefacts to the resulting segmentation. An example can be found in Fig~\ref{fig:regisration} $d$ and $f$. In this section, we introduce shape prior knowledge through atlas propagation to overcome such artefacts in SSLLN-LR segmentation. In Fig~\ref{fig:regisration}, we outline the shape refinement framework, including initial affine alignment, atlas selection, deformable registration and label fusion. The framework involves using a cohort of high-resolution (HR) atlases produced from SSLLN-HR, each of which consists of an HR CMR volume ($1.25 \times 1.25 \times 2.0\;\rm{mm}$), and its corresponding landmarks and segmentation labels. Next, we detail the framework.

Due to individual differences, the scanned heart often shows marked variations in size, pose and shape (as shown in Fig~\ref{fig:regisration} $a$ and $d$). This poses difficulty for existing image registration algorithms due to their non-convex nature. For this, the landmarks detected from SSLLN-HR and -LR were used to initialise the subsequent non-rigid algorithm between target and each atlas, which is similar to \cite{grbic2012complete,duan2018combining}. An affine transformation with 12 degrees of freedom was first computed between the target landmarks (predicted by SSLLN-LR) and the atlas landmarks (predicted by SSLLN-HR). In addition to initialising the non-rigid image registration, the resulting affine transformations were used to warp segmentations in all atlases to the target space for atlas selection. According to the normalised mutual information (NMI) scores between the target segmentation and each of affinely warped atlas segmentations, $L$ most similar atlases can be selected to save registration time and to remove dissimilar atlases for label fusion.

Since the correspondences of structures across both target and atlas volumes are explicitly encoded in their segmentations, we only use segmentations for the following non-rigid registration. Let $S$ and $l_n$ ($n=1,...,L$) be the SSLLN-LR segmentation and the $n$th atlas segmentation, respectively. Let $P_{S,{l_n}}(i,j)$ be the joint probability of labels $i$ and $j$ in $S$ and $l_n$, respectively. It is estimated as the number of voxels with label $i$ in $S$ and label $j$ in $l_n$ divided by the total number of voxels in the overlap region of both segmentations. We then maximise the overlap of structures denoted by the same label in both $S$ and ${l_n}$ by minimising the following objective function 
\begin{equation} \label{eq:LC}
\Phi _n^* = \arg \min {\cal C}\left( {{S},{{{l_n}}}({\Phi _n})} \right)
\end{equation}
where $\Phi_n$ is the transformation between $S$ and ${l_n}$, which is modelled by a free-form deformation (FFD) based on B-splines \cite{rueckert1999nonrigid}. ${\cal C}( {{S},{{{l_n}}}})=\sum_{i=1}^{N_r} P_{S,{l_n}}(i,i)$, representing the label consistency \cite{frangi2002automatic}. ${\cal C}$ in (\ref{eq:LC}) is a similarity measure of how many labels, of all the labels in the atlas segmentation, are correctly mapped into the target segmentation. With the affine transformation as initialisation, a multi-scale gradient descent was then used to minimise the objective function (\ref{eq:LC}). After the optimal $\Phi^*_n$ is found, the segmentations and volumes in the $n$th atlas are warped to the target space. The process is repeated until $n=L$.

Lastly, we perform non-local label fusion to generate an accurate and smooth bi-ventricular model $\tilde S$ for the imperfect SSLLN-LR segmentation $S$. Let us first denote the warped atlas volumes and segmentations as $\{(f_{n},l^{\prime}_{n})|n=1,...,L\}$, respectively. Here, $n$ denotes the warped atlas index and $L$ is the number of selected atlases. For each voxel $x$ in the target LR volume $f$, a patch $f_x$ centred at $x$ can be constructed. The aim of the label fusion task is to determine the label at $x$ in $f$ using $\{(f_{n},l^{\prime}_{n})|n=1,...,L\}$. For each voxel $x$ in $f_{n}$, we define $\{(f_{n,y},l_{n,y})|n=1,...,L, y \in {\cal N}(x)\}$, where $y$ denotes a voxel in the search window ${\cal N}(x)$, $f_{n,y}$ denotes the patch centred at voxel $y$ in the $n$th warped atlas, and $l_{n,y}$ denotes the corresponding label for voxel $y$. The resulting label at voxel $x$ in the target volume $f$ can be calculated as
\begin{equation} \label{eq:labelFusion}
{S_x} = \mathop {\arg \max }\limits_{k = 1,..,{N_r}} \smashoperator{\sum\limits_{n}} \smashoperator{\sum\limits_{\;\;y \in {\cal N}(x)}} { {{e^{ - \frac{{\| {{f_x} - {f_{n,y}}}\|_F^2}}{h}}} \cdot {\delta _{{l_{n,y}},k}}} } 
\end{equation}
where $h$ denotes the bandwidth for the Gaussian kernel function and ${\delta _{{l_{n,y}},k}}$ denotes the Kronecker delta, which is equal to one when $l_{n,y}=k$ and equal to zero otherwise. The equation (\ref{eq:labelFusion}) can be understood as a form of weighted voting, where each of the patches from each of the atlases contributes a vote for the label. It is a non-local method because it uses patch similarity formulation (i.e. Gaussian kernel function), which is inspired by the non-local methods \cite{buades2005non,duan2015fast,lu2019graph}. It has been shown in \cite{bai2015multi} that, in a Bayesian framework, (\ref{eq:labelFusion}) is essentially a weighted $K$ nearest neighbours (\textit{KNN}) classifier, which determines the label by maximum likelihood estimation. By aggregating high-resolution atlas shapes in this way, an explicit anatomical shape prior can be inferred. The artefacts in the SSLLN-LR segmentation can thus be resolved, as shown in Fig.~\ref{fig:regisration} $j$. 
\section{Experiments}
\label{experiments}
In this section, we cover extensive experiments to evaluate (both qualitatively and quantitatively) the performance of the proposed pipeline on short-axis CMR volumetric images. Dice index and Hausdorff distance \cite{bai2017human} were employed for evaluating segmentation accuracy. Dice varies from 0-1, with high values corresponding to a better results. The Hausdorff distance is computed on an open-ended scale, with smaller values implying a better match. We also validate the performance using clinical measures (ventricular volume and mass) derived from the segmentations. In the following experiments, each component in the pipeline is studied separately.

\subsection{Clinical datasets}
\label{clinicaldataset}
\textit{\textbf{UK Digital Heart Project Dataset}}: This dataset\footnote{\href{https://digital-heart.org/}{https://digital-heart.org/}} (henceforth referred to as Dataset 1) is composed of 1831 cine HR CMR volumetric images from healthy volunteers, with corresponding dense segmentation annotations at the end-diastolic (ED) and end-systolic (ES) frames.  {The ground-truth segmentation labels were manually annotated by a pair of clinical experts working together, and each volume was only annotated by one expert at a time}. For each volume at ED, 6 landmarks, as shown in Fig~\ref{fig:exampleFig} middle, were manually annotated by a clinician {(inter-user 1)}. The raw volumes were derived from healthy subjects, scanned at Hammersmith Hospital, Imperial College London using a 3D cine balanced steady-state free precession (b-SSFP) sequence \cite{de2014population} and has a resolution of $1.25 \times 1.25 \times 2 \;\rm{mm}$. As introduced in Section \ref{overview}, HR imaging technique does not produce cardiac artefacts which are often seen in LR imaging acquisition \cite{petersen2015uk}. 

\textit{\textbf{Pulmonary Hypertension Dataset}}: This dataset (henceforth referred to as Dataset 2) was acquired at Hammersmith Hospital National Pulmonary Hypertension Centre, and composed of 649 subjects with pulmonary hypertension (PH) - a cardiovascular disease characterised by changes in bi-ventricular volume and geometry. PH subjects often have breathing difficulties, therefore HR imaging was impractical for the majority of patients in this cohort due to the relatively long breath-hold time required. Within the cohort, 629 of the 649 patients were scanned using conventional LR image acquisition, and this manner of image acquisition (over multiple short breath-holds) often leads to lower-resolution volumes and inter-slice shift artefacts. In contrast, the remaining 20 subjects managed to perform a single breath-hold, and therefore HR volumes could be acquired for these subjects. Coupled with these HR volumes, LR volumes were also acquired during scanning, forming 20 pairs of LR and HR cine CMR volumes. The resolutions for LR and HR volumes are $1.38 \times 1.38 \times 10 \;\rm{mm}$ and $1.25 \times 1.25 \times 2 \;\rm{mm}$, respectively. For all 649 subjects, the manual ground-truth segmentation labels at ED and ES were generated, and 6 landmarks at ED were also annotated.

\subsection{Preprocessing and augmentation}
\label{circle}
\textbf{{Preprocessing}}: Image preprocessing was carried out to ensure: 1) the size of each volumetric image fits the network architecture; 2) the intensity distribution of each volume was in a comparable range so that each input could be treated equally importantly. As such, each of the HR volumes in Dataset 1 was reshaped to common dimensions of $192 \times 192 \times 80$ with zero-padding if necessary, while each of LR volumes in Dataset 2 was interpolated to $1.25 \times 1.25 \times 2$ mm and then reshaped to $192 \times 192 \times 80$. For the best visual effect, the figures shown in experiments may be cropped manually. However, no ROI detection algorithm (for localisation of the heart) was used in image preprocessing. The intensity redistribution processes for both HR and LR volumes are the same. After reshaping, we first clipped the extreme voxel values (i.e. outliers) in each HR/LR volume. We defined outliers as voxel values lying outside of the 1st to 99th percentile range of original intensity values. Finally, the resulting voxel intensities of each volume were scaled to the $[0, 1]$ range.

\textbf{{Parameter selection}}: The following parameters were utilised for the experiments in this study: For training the network, each run was carried out for 50 epochs, with batch size of 8 volumes, learning rate of 0.001 and Adam stochastic gradient descent for optimisation. The weight coefficients $\alpha$, $\beta$ and $\gamma$ in (\ref{eq:SDLloss}) are empirically set to 0.8, 0.2 and $5 \times 10 ^ {-5}$, respectively. The small positive value in the Dice loss (\ref{eq:Dloss}) is set to $1 \times 10^{-8}$. According to \cite{baumgartner2017exploration} the exact network architecture only plays a minor role in improving segmentation accuracy. Therefore, the network architecture, as shown in Fig~\ref{fig:network}, was used without significant modification. For the non-local label fusion (\ref{eq:labelFusion}), we used a value of 10 for the bandwidth parameter $h$, voxel dimensions $7 \times 7 \times 1$ for the patch window size and $7 \times 7 \times 3$ for the search window size. For more details on parameter tuning in (\ref{eq:labelFusion}), we refer the reader to \cite{bai2015multi}. Finally, $L=5$ atlases were used for label fusion. Using the parameter settings outlined above, we found the pipeline performed very well for our experiments, indicating its robustness to parameters tuning.

\textbf{{Augmentation}}: Since our network takes volumetric images as inputs, we performed 3D data augmentation on-the-fly during training. At each iteration, augmentation included rescaling of voxel intensities in the input volume, and a 3D random affine transformation of the volume and corresponding labels and landmarks. For simplicity, the affine transformation only involved in-plane translation, isotropic scaling and rotation along one random direction ($x$-, $y$- or $z$-axis) at the central voxel of the volume. Neither shearing nor volume flipping was used. Data augmentation enables the network to see a large and diverse array of inputs by the end of training, and was implemented using the \textit{SimpleITK} library in \textit{Python}. With an Nvidia Titan XP GPU, training (50 epochs) took approximately 20 and 10 hours for Datasets 1 and 2, respectively. For inference, segmentation (without shape refinement) of an HR/LR volume for a single subject at ED took $<1s$. 

\subsection{Segmentation of high-resolution volumes}
\label{HRExp}
First, we conducted experiments using Dataset 1, which includes 1831 HR CMR volumes. We randomly split the dataset into {two disjoint subsets of 1000/831. The first subset was used to train SSLLN-HR, and the second subset was used for testing the accuracy of segmentation and landmark localisation, respectively}. During training, we only used ED instances (volumes, landmarks and segmentation labels). Note that the proposed SSLLN-HR is a multi-task network that simultaneously outputs labels and landmarks. Next we segmented a cardiac volume into 5 regions: the left ventricular cavity (LVC), right ventricular cavity (RVC), left ventricular wall (LVW), right ventricular wall (RVW) and background. Our method is the first one capable of producing a full HR bi-ventricular segmentation (LVC+LVW+RVC+RVW) in 3D.
\begin{figure}[h!]
\centering  
{\includegraphics[width=0.48\textwidth]{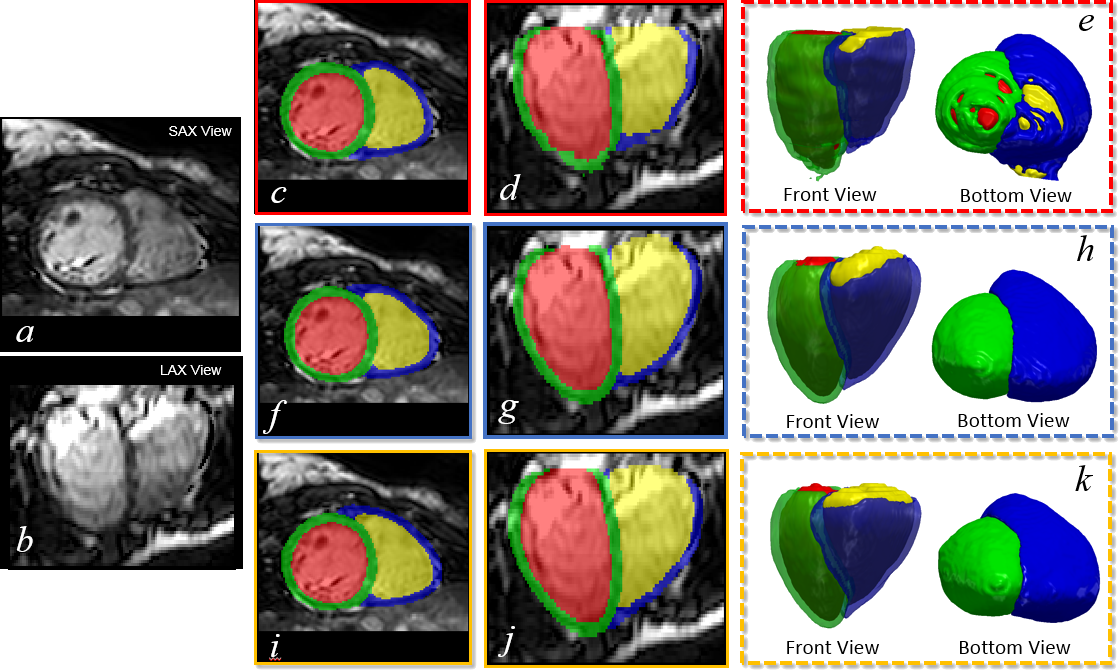}}\\
\vspace{-10pt}
\caption{Visual comparison of segmentation results by 2D slice-by-slice FCN, 3D FCN and SSLLN-HR. $a$ and $b$: two views of a high-resolution volume; $c$, $d$ and $e$: results by 2D FCN; $f$, $g$ and $h$: results by 3D FCN; $i$, $j$ and $k$: SSLLN-HR. SAX and LAX denote short-axis and long-axis, respectively.}
\label{fig:2D3D}
\end{figure}

{In Fig~\ref{fig:2D3D}, we compare SSLLN-HR with two baseline methods for segmentation. The first one is the 2D FCN proposed in \cite{bai2017human}, where the network\footnote{Code is publicly available at \href{https://github.com/baiwenjia/ukbb_cardiac}{https://github.com/baiwenjia/ukbb$\_$cardiac}} was trained using each short-axis slice in the volume separately. The second one is the 3D FCN, whose architecture is similar as in Fig~\ref{fig:network}. To make the 3D FCN fit GPU memory, we halved the number of activation maps in each layer (excluding last one) and cropped the original image to a size of $112 \times 112 \times 64$. To focus exclusively on segmentation accuracy, we removed the landmark localisation activation maps in the last layer of the 3D FCN. As Fig~\ref{fig:2D3D} shows, 2D FCN produces a jagged appearance as shown in the long-axis view image Fig~\ref{fig:2D3D} $d$, and there are `cracks' in the corresponding 3D model as shown in Fig~\ref{fig:2D3D} $e$. This problem is due to the fact that the 2D method does not consider 3D context of the volumetric image, leading to a lack of spatial consistency between segmented slices. In contrast, both SSLLN-HR and 3D FCN account for the spatial consistency between slices, enabling smooth results. Visually, SSLLN-HR is comparable to 3D FCN. However, SSLLN-HR is less memory demanding and therefore can be directly implemented on non-cropped volumes with a faster training speed.}

\begin{table}[h!]\centering
\caption{Dice index and Hausdorff distance derived from 2D FCN, 3D FCN, and SSLLN-HR for segmenting 831 high-resolution short-axis volumetric images. The mean $\pm$ standard deviation are reported.}
\vspace{-5pt}
\resizebox{\columnwidth}{!}
{\begin{tabular}{ccccccccc}
\toprule
& \multicolumn{3}{c}{Dice Index ($\%$)}     &     & \multicolumn{3}{c}{Hausdorff Dist. (mm)}            \\
\cmidrule{2-4} \cmidrule{6-8} 
& 2D FCN       &   3D FCN        & SSLLN-HR & & 2D FCN            &     3D FCN      & SSLLN-HR\\ 
\midrule
LVC & 0.950$\pm$0.022  & \textbf{0.963$\pm$0.010}     & 0.962$\pm$0.015 		&	& 2.584$\pm$1.108		  & \textbf{2.037$\pm$0.413}      & {2.203$\pm$0.922}&\\ 
LVW & 0.836$\pm$0.060  & \textbf{0.888$\pm$0.024}    & 0.873$\pm$0.034   		&   & 3.927$\pm$1.712		  & \textbf{3.028$\pm$1.062}      & {3.242$\pm$0.992}&\\ 
RVC & 0.887$\pm$0.061  & {0.917$\pm$0.025}    & \textbf{0.929$\pm$0.026}      	&	& 6.614$\pm$4.032         & {4.748$\pm$1.253}      & \textbf{4.171$\pm$1.527}&\\
RVW & 0.633$\pm$0.132  & 0.732$\pm$0.073      & \textbf{0.755$\pm$0.068}    	&	& {8.252$\pm$3.644}  	  & 6.184$\pm$1.403	       & \textbf{5.996$\pm$1.424}&\\ 
\bottomrule
\end{tabular}}
\label{tb:2D3D_}
\end{table}

\begin{figure}[h!]
\centering  
{\includegraphics[width=0.5\textwidth]{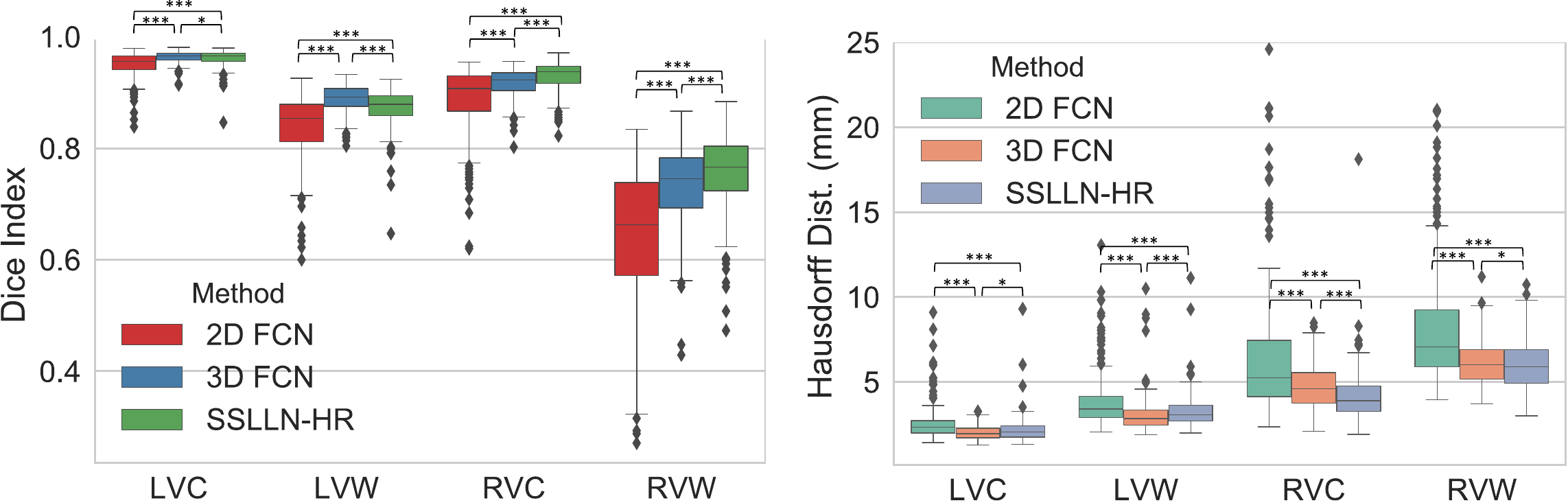}}\\
\vspace{-5pt}
\caption{{Boxplot comparison of segmentation accuracy between 2D FCN, 3D FCN and SSLLN-HR on 831 high-resolution short-axis volumetric images. The symbol `***' denotes $p$ $\ll$ 0.001, and `*' denotes $p$ $<$ 0.1.}}
\label{fig:boxplot}
\end{figure}

{Table~\ref{tb:2D3D_} provides a summary of quantitative comparisons between 2D FCN, 3D FCN and SSLLN-HR, with statistics derived from 831 subjects. Statistical significance of the observed differences in the evaluation metrics (Dice index and Hausdorff distance) between each pair of methods is assessed via the Wilcoxon signed-rank test. The results in the table demonstrate the high consistency between automated and manual segmentations. In terms of Dice and Hausdorff distance, SSLLN-HR and 3D FCN outperformed 2D FCN, and SSLLN-HR achieved comparable performance to 3D FCN. Of note, all three methods achieved a relative low Dice score on the RVW anatomy. This is due to the thinness of RVW and the fact that the Dice index is more sensitive to errors in this structure. In Fig~\ref{fig:boxplot}, boxplots visually depicting the results of Table are presented. As these plots show, the 2D method produced large variation across different segmentations for the four anatomies, resulting in a inferior accuracy than the 2.5D and 3D methods. SSLLN-SR achieved similar results to 3D FCN, with the segmentation accuracy of RVC and RVW slightly higher than that of 3D FCN.}

\begin{figure}[h!]
\vspace{-5pt}
\centering  
{\includegraphics[width=0.48\textwidth]{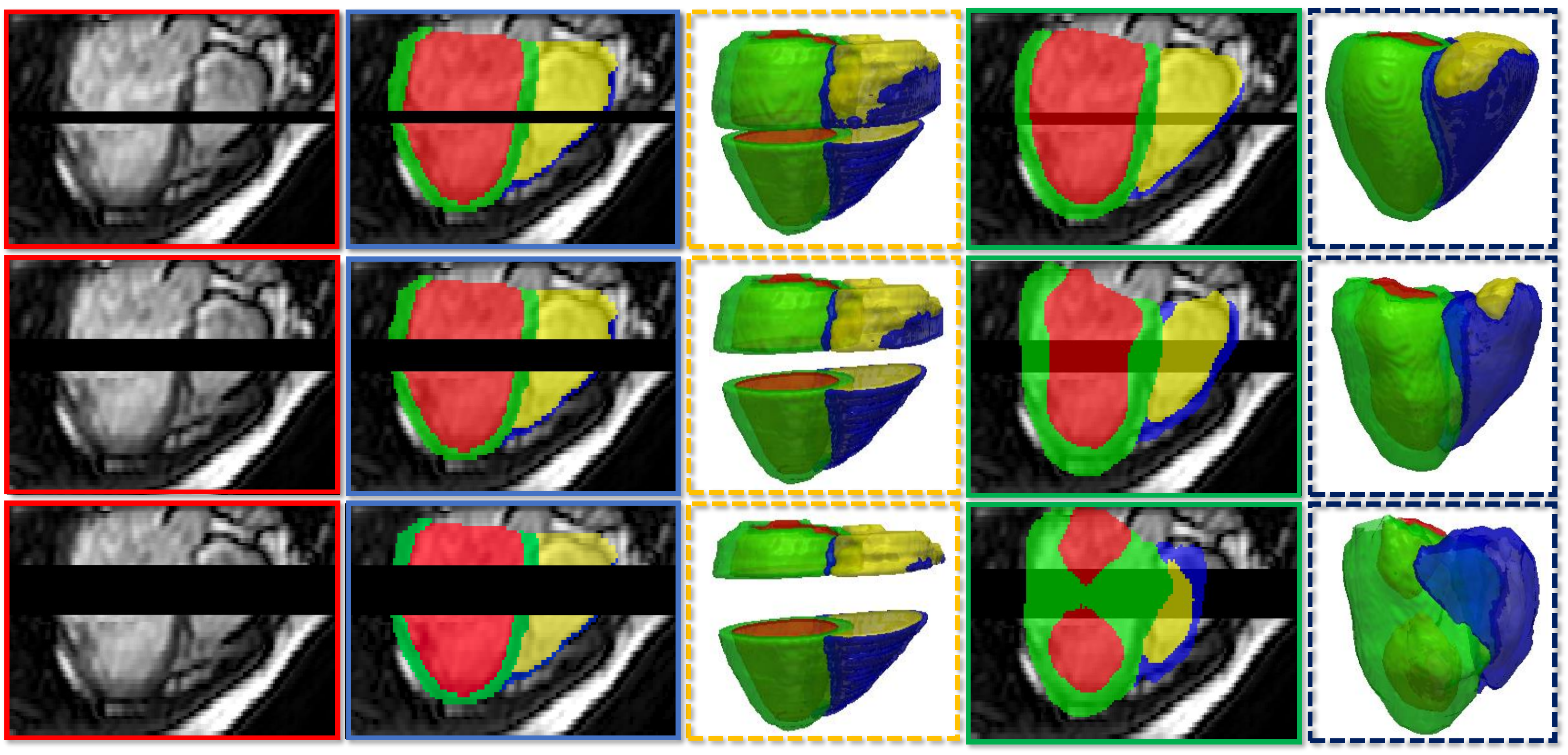}}\\
\vspace{-5pt}
\caption{Testing 3D spatial consistency of the 2D FCN and SSLLN-HR methods. 1st column: target segmentation volumes with zero-filled gaps of different sizes; 2nd and 3rd columns: 2D FCN results; 4th and 5th columns: SSLLN-HR results. }
\label{fig:gap}
\end{figure}
In Fig~\ref{fig:gap}, we further compare the proposed SSLLN-HR with the 2D FCN. We selected batches of $k$ consecutive short-axis slices in a volumetric image, with multiple settings of $k$ (=5,13, and 20). In each case, we set intensities in the selected slices to zero, as shown in the 1st column. The two methods under comparison were then applied to these partially zero-filled volumes, and the results are given in 2nd-5th columns. As is evident, 2D FCN fails to segment these zero-filled slices, thus leaving gaps in the resulting 3D segmentations. In contrast, SSLLN-HR demonstrates robustness to missing slices and has the capability of `inpainting' these gap regions. However, as the gap (number of zero-filled slices) increases (from $k$=5 to $k$=20), the segmentation performance becomes worse. These results further illustrate that the proposed network retains 3D spatial consistency, which the 2D FCN is unable to achieve. Our method thus outperforms the 2D approach in this regard.

\subsection{Landmark localisation}
\label{Landmark}
{To enable automatic alignment for subsequent non-rigid registration, we also predicted landmark locations (together with segmentation) for each input volume using SSLLN-HR. Same as above, we used the split subsets 1000/831 for training and testing. Note that SSLLN-HR was trained with manual landmarks carried out by inter-user 1 on each of the 1000 subjects. For the 831 unseen test subjects, the automatically detected landmarks were compared with the manual ones from inter-user 1 using the point-to-point Euclidean distance. Also, to study inter-user variability of landmarking, a second expert (inter-use 2) was recruited to manually annotate landmarks for each of 831 test subjects. The annotations were then compared with those of inter-user 1.}

\begin{figure}[h!]
\centering  
{\includegraphics[width=0.5\textwidth]{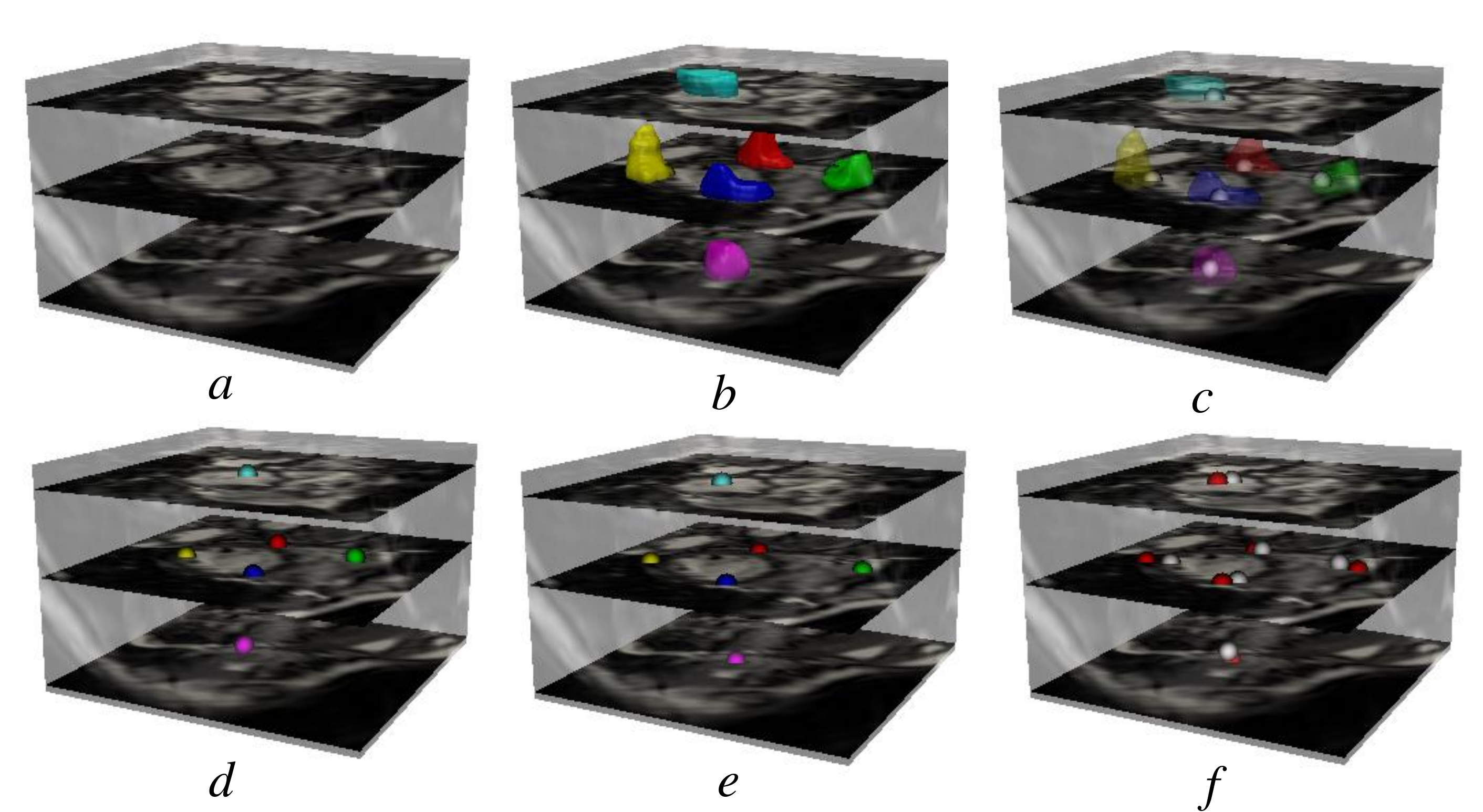}}\\
\vspace{-5pt}
\caption{Landmark localisation using the proposed network. $a$: input volume; $b$: landmarks detected by SSLLN-HR directly; $c$: single-voxel landmark extraction from each clustered landmark in $b$; $d$: colour coded single-voxel landmarks; $e$: ground-truth landmarks annotated by inter-user 1; $f$: superimposed ground-truth (red) and automated (white) landmarks.}
\label{fig:ldmks}
\end{figure}

Fig~\ref{fig:ldmks} first shows a visual comparison of automated and manual (inter-user 1) landmarks. Fig~\ref{fig:ldmks} $b$ shows the landmark locations predicted by our SSLLN-HR. As is evident, each landmark is represented by a few locally clustered voxels. The central gravity (represented by a single voxel) of each landmark in Fig~\ref{fig:ldmks} $b$ can be computed by averaging the positions of the voxels forming the true landmark. The corresponding results are shown in Fig~\ref{fig:ldmks} $c$, where the two type of landmarks are superimposed. The respective colour-coded single-voxel landmarks are shown in Fig~\ref{fig:ldmks} $d$, which were used for initial point-to-point affine registration, as shown in Fig~\ref{fig:regisration}. In Fig~\ref{fig:ldmks} $f$, we superimposed the automated detected landmarks and manual landmarks (Fig~\ref{fig:ldmks} $e$). As can be seen, Fig~\ref{fig:ldmks} $f$ demonstrates very good consistency between the automated and manual landmarks. 
\vspace{-5pt}

{In Table~\ref{tb:p2p1}, we compare the landmark localisation errors between automated and manual methods, as well as between the two manual methods on 831 test volumes. Using inter-user 1 as a baseline for comparison, we observe that the SSLLN-HR detections are more accurate than the annotations of inter-user 2. The point-to-point distance errors between SSLLN-HR and inter-user 1 vary only from 3.67$\pm$3.20 mm for Landmark-I to 8.18$\pm$6.91 mm for Landmark-II. In contrast, the errors between the two inter-users vary from 5.61$\pm$2.62 mm for Landmark-V to 17.4$\pm$9.27 mm for Landmark-II. This confirms that computer-human difference can be smaller than human-human difference.}
\vspace{-15pt}
\begin{table}[h!]
\centering
\caption{{Point-to-point (P2P) distance error statistics in landmark localisation over 831 volumes. The second column shows the errors between automated (SSLLN-HR) and manual (inter-user 1) landmark localisations. The third column shows the errors between two manual (inter-user 1 and 2) annotations. The mean $\pm$ standard deviation in mm are reported. The description of the 6 landmarks is given in Fig~\ref{fig:exampleFig}.}}
{\begin{tabular}{lcccc}
\toprule
Region & {Auto vs Man 1} & {Man 1 vs Man 2} & {$p$-value}  \\  
\midrule
Landmark-I (blue)    & 3.67$\pm$3.20  & 9.16$\pm$4.37  & $p\ll$0.001     \\ 
Landmark-II (green)  & 8.18$\pm$6.91  & 17.4$\pm$9.27  & $p\ll$0.001     \\ 
Landmark-III (red)   & 3.99$\pm$3.54  & 9.69$\pm$5.39  & $p\ll$0.001    \\ 
Landmark-IV (yellow) & 5.77$\pm$4.15  & 11.7$\pm$5.87  & $p\ll$0.001     \\ 
Landmark-V  (pink)   & 3.86$\pm$2.74  & 5.61$\pm$2.62  & $p\ll$0.001     \\ 
Landmark-VI (cyan)   & 5.15$\pm$2.82  & 14.6$\pm$5.44  & $p\ll$0.001    \\ 
\bottomrule
\end{tabular}}
\label{tb:p2p1}
\end{table}
\vspace{-5pt}
\begin{figure}[h!]
\centering  
{\includegraphics[width=0.5\textwidth]{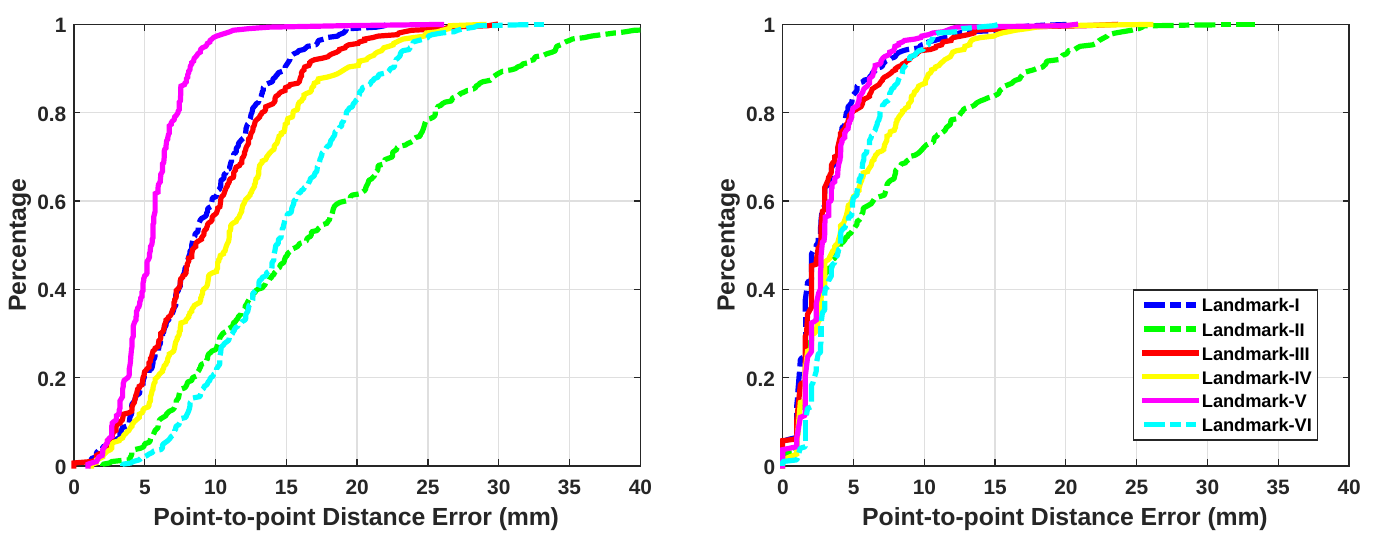}}\\
\vspace{-5pt}
\caption{{Cumulative error distribution curves of landmark localisation errors. The left curves are derived from manual landmarks of inter-user 1 and inter-user 2, and the right curves are plotted based on automated (SSLLN-HR) and manual (inter-user 1) landmark localisations. }}
\label{fig:cdf}
\end{figure}

{Fig~\ref{fig:cdf} provides a simple visualisation of the relative error distribution in the test sample of 831 volumes. The two plots show the cumulative distribution of point-to-point distance error for each landmark. As can be seen, the curves in the right are more clustered and stacked vertically than those in the left, indicating superior accuracy of landmark localisations by SSLLN-HR. For example, from the right plot we see that for all landmarks, about 92$\%$ of test volumes had point-to-point distance error of $<$20 mm. In contrast, only 60$\%$ of test volumes reached point-to-point distance error of $\sim$20 mm, as shown in the left plot.}
\vspace{-5pt}
\begin{figure}[h!]
\centering  
{\includegraphics[width=0.48\textwidth]{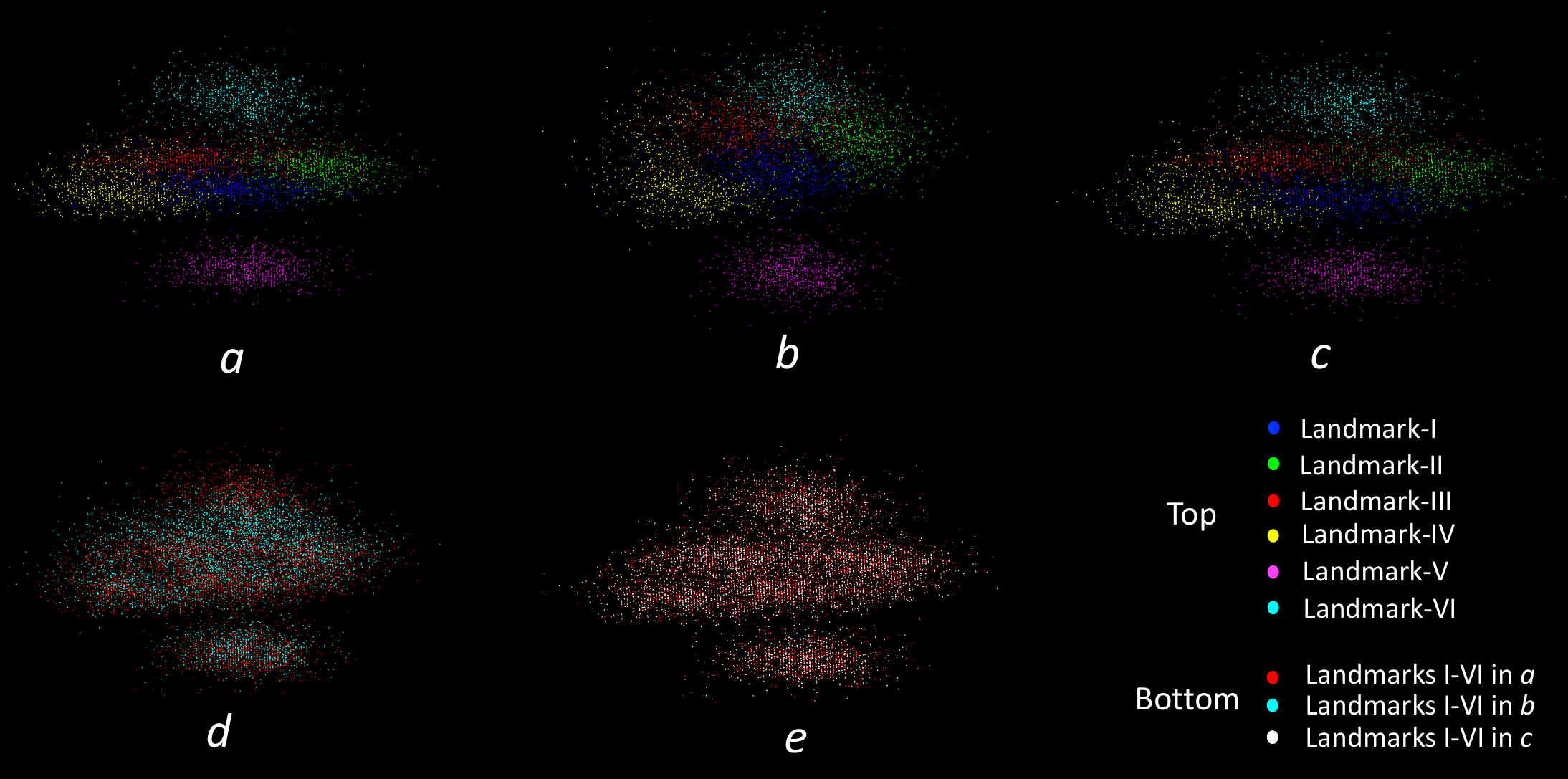}}\\
\vspace{-5pt}
\caption{{Visualisation of landmarks in 3D. $a$: manual landmarks by inter-user 1; $b$: manual landmarks by inter-user 2; $c$: landmarks localised by the network, trained on the manual annotations of inter-user 1; $d$: superimposed $a$ and $b$; $e$: superimposed $a$ and $c$. It is evident that $a$ and $c$ overlap to a greater degree than $b$ and $c$.}}
\label{fig:allLdmks}
\end{figure}

\begin{figure*}[h!]
\vspace{-5pt}
\centering  
{\includegraphics[width=0.99\textwidth]{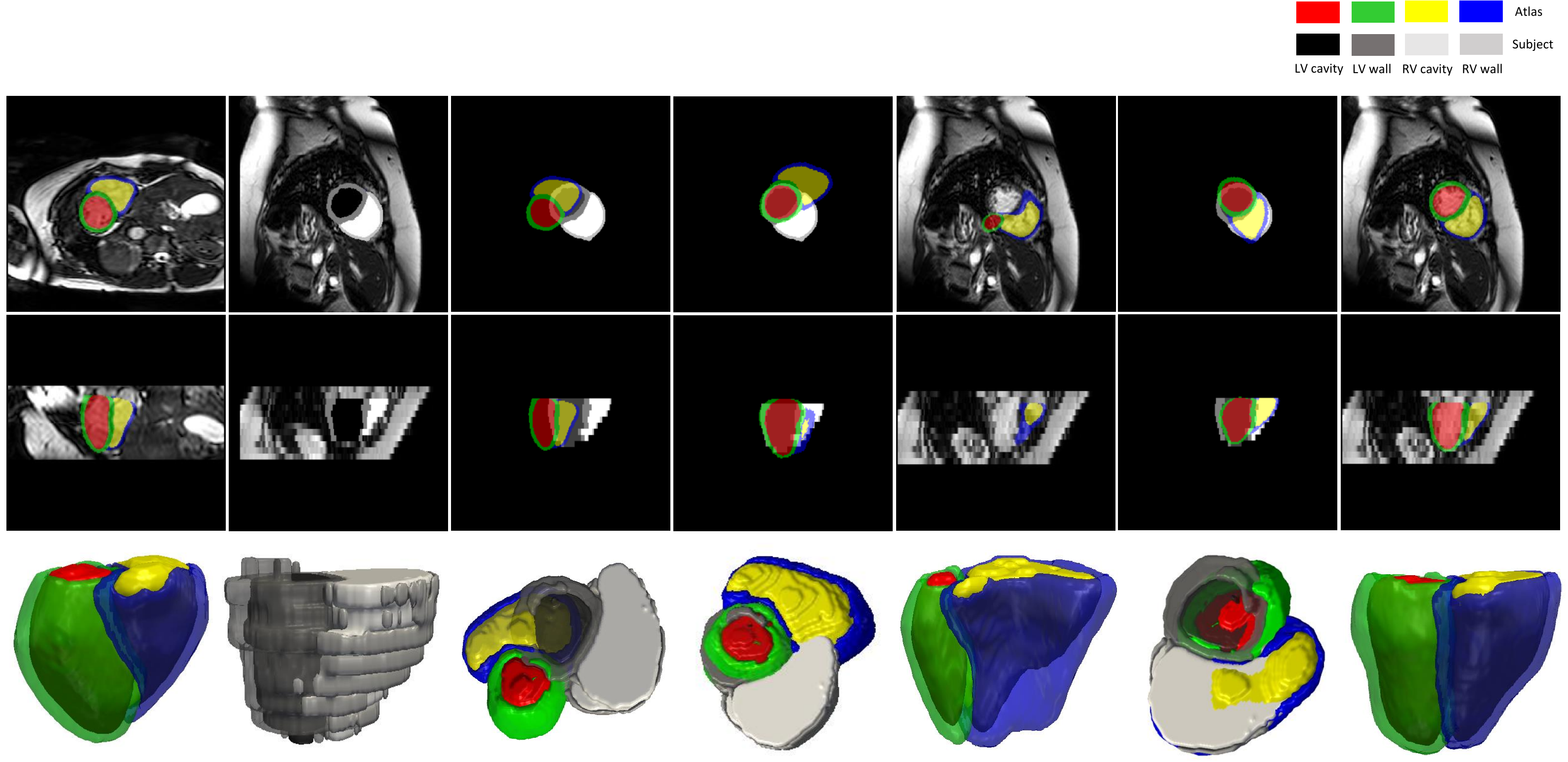}}\\
\vspace{-5pt}
\caption{Impact of using landmarks in the proposed pipeline. 1st column: SSLLN-HR segmentation of an HR atlas volume; 2nd column: SSLLN-LR segmentation of an LR volume; 3rd column: superimposed SSLLN-HR and SSLLN-LR segmentation labels; 4th column: SSLLN-HR segmentation affinely warped to the SSLLN-LR segmentation based on their labels; 5th column: final SSLLN-LR segmentation refined by the non-rigid registration initialised with the label-based affine transform; 6th column: SSLLN-HR segmentation affinely warped to the SSLLN-LR segmentation based on their landmarks localised by the network in Fig~\ref{fig:network}; 7th column: final SSLLN-LR segmentation refined by the non-rigid registration initialised with the landmark-based affine transform. The 1st, 2nd and 3rd rows respectively show the short-axis view, long-axis view and corresponding 3D visualisation of segmentation.}
\label{fig:impact}
\end{figure*}

{Fig~\ref{fig:allLdmks} provides a 3D visualisation of landmarks for all the 831 volumes. These landmarks were acquired from inter-user 1, inter-user 2 and SSLLN-HR. This figure further illustrates that SSLLN-HR, trained from manual annotations of a human, excellently matches the performance of that human on an unseen test set. On the other hand, the discrepancy between human-human performance could be very large. Fig~\ref{fig:allLdmks}, together with Fig~\ref{fig:ldmks}, Table~\ref{tb:p2p1} and Fig~\ref{fig:cdf}, provide an ample evidence that the proposed SSLLN-HR has the capability of detecting landmarks robustly and accurately, and that it tends to produce less variability in predictions relative to variability among human experts. }

\subsection{Impact of landmarks}
\label{simluated}
{In this section, we show that landmarks localisation is a necessary step in our pipeline. In Fig~\ref{fig:impact}, we compare the SSLLN-LR segmentation results refined by the non-rigid deformation with different initialisations of affine transformation. As shown in the 5th column of Fig~\ref{fig:impact}, the non-rigid refinement failed completely if the affine transform is initialised from the tissue classes. In contrast, initialising it directly on the landmarks resulted in an accurate refinement, as shown in the last column of Fig~\ref{fig:impact}. We propose two reasons for this observation: 1) the six anatomical landmarks defined in the study effectively reflect the underlying pose, size and shape of a heart. As such, warping a heart with landmark-based affine transformation produces a very robust initialisation for the subsequent non-rigid registration; 2) Computing an affine transformation from a pair of landmarks is a convex least squares problem, a unique solution to which exits. In contrast, initialising an affine transform directly on the tissue classes is a non-convex problem. As such, the warped result is sometimes sub-optimal, which may negatively impact the non-rigid registration and increase uncertainty of the registration method. Moreover, label-based affine registration is much more computational expensive than landmark-based affine registration as it needs to deal with millions of voxels in the 3D volumes. In the \textit{IRTK} implementation\footnote{Code is publicly available at \href{https://github.com/BioMedIA/IRTK}{https://github.com/BioMedIA/IRTK}}, it took $\sim$0.005s to compute an affine transformation on a pair of landmarks, whilst it took $\sim$5s to perform an affine registration using the 3D segmentation labels with size $256 \times 256 \times 56$.}

{We note that it may also be possible to detect landmarks automatically from segmentation labels. In this case, the accuracy of landmarks will be conditioned on the accuracy of segmentation. On the other hand, it may not be straightforward to determine which landmarks should be detected from segmentation labels for robust registration. As such, directly detecting the six landmarks defined in the study using the proposed network is neater and better.}

\subsection{Experiments on simulated low-resolution volumes}
\label{simluated}
\begin{figure}[h]
\vspace{-5pt}
\centering  
{\includegraphics[width=0.5\textwidth]{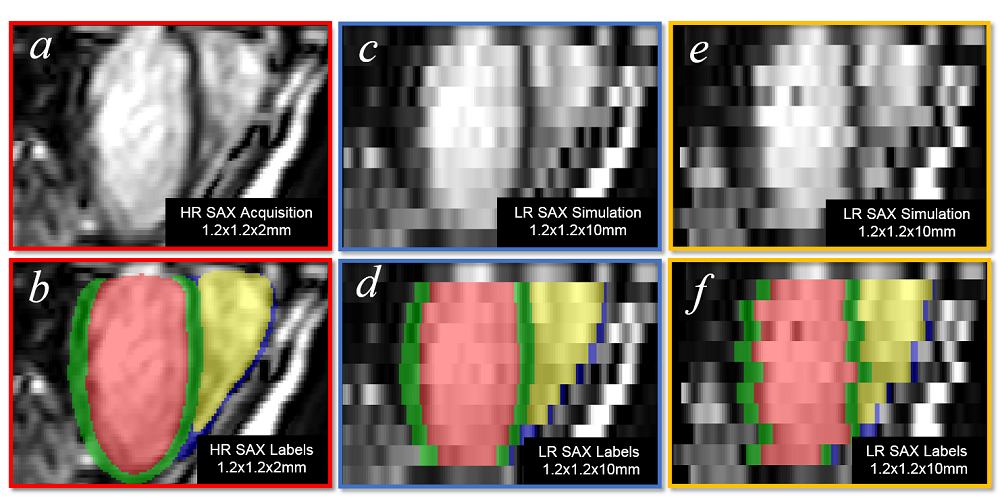}}\\
\vspace{-5pt}
\caption{Simulating cardiac artefacts in real scenarios. 1st column: artefact-free high-resolution cardiac volume and ground-truth labels. 2nd column: downsampled versions of volumes in the 1st column. 3rd column: inter-slice shift is added to the downsampled volumes in the 2nd column.}
\label{fig:visual}
\end{figure}

To quantitatively assess the performance of SSLLN-LR and shape refinement (SSLLN-LR+SR) in the pipeline (bottom path in Fig~\ref{fig:flowchart}), we developed a method to simulate different types of artefacts seen in LR cardiac volumes. Specifically, in Fig~\ref{fig:visual} an HR volume and its manual segmentation were first downsampled from $1.25 \times 1.25 \times 2 \;\rm{mm}$ to $1.25 \times 1.25 \times 10 \;\rm{mm}$, as shown in the 1st and 2nd columns. The downsampling produces a staircase artefact due to reduction in long-axis resolution. Moreover, the segmentation (Fig~\ref{fig:visual} $d$) around the apical region is now incomplete due to the lack of coverage of the whole heart. We further simulated inter-slice shift artefact by randomly translating each 2D short-axis slice horizontally. This step produced misalignment in the cardiac volume and its segmentation, as shown in the 3rd column. 

Next, for training the SSLLN-LR, the LR volume Fig~\ref{fig:visual} $e$ and its segmentation Fig~\ref{fig:visual} $f$ were used as inputs. Note that our method is capable of producing an HR smooth segmentation model even from misaligned inputs such as the example in Fig~\ref{fig:visual} $f$. Since we have the smooth ground truth Fig~\ref{fig:visual} $b$ for the simulated Fig~\ref{fig:visual} $e$, we can quantitatively assess the ability of our method to recover the original smooth shape. For these simulation experiments, we split Dataset 1 into subsets (1000/600/231). The first two subsets were corrupted with the simulated artefacts described above, which were used for training the SSLLN-LR and testing the proposed shape refinement (SC) component of the pipeline. The HR atlas shapes (segmented by SSLLN-HR) in the last cohort ($n$=231) were used to refine SSLLN-LR segmentations. 

Here we highlight three reasons why we used SSLLN-HR network results as a reference atlas set for shape-refinement: 1) our SSLLN-HR is able to produce results that are very similar to the corresponding ground truth, as confirmed from Section \ref{HRExp} and \ref{Landmark}; 2) Once the SSLLN-HR is trained, it can be readily deployed on an external dataset (where HR atlases are not available) to create new HR atlases so as to facilitate the running of our pipeline; 3) The atlas set can be enriched by adding more SSLLN-HR results, which will increase the possibility to select better atlases for the sequential registration-based refinement.

In Table~\ref{tb:2D3D}, we compare the Dice index and Hausdorff distance between the SSLLN-HR and SSLLN-LR+SR results. SSLLN-HR was directly evaluated on 600 artefact-free HR volumes at ED in Section~\ref{HRExp}, while SSLN-LR+SC was tested on the 600 corresponding simulated LR volumes where cardiac artefacts exist, as shown in Fig~\ref{fig:visual}. Although SSLLN-HR performs better than SSLLN-LR+SR, the performance gap between two approaches is minor. For LVC, LVW and RVC, the Dice index of SSLLN-HR is only about 0.2 higher than that of SSLLN-LR+SR. The Hausdorff distance of SSLLN-HR is about 0.5 mm smaller than that of SSLLN-LR+SR for all 4 regions. Again due to the thin structure of RVW, the mean Dice values of the two methods are relatively low: 0.662 and 0.557, respectively. This table shows that SSLLN-LR+SR achieves good segmentation results for imperfect LR input volumes, and the results are comparable to direct segmentation of artefact-free HR results.
\begin{table}[h!]
\centering
\caption{Comparison of Dice index and Hausdorff distance between SSLLN-HR and SSLLN-LR+SR (shape refinement). SSLLN-HR was validated on 600 high-resolution short-axis volumetric images from Dataset 1, whilst SSLLN-LR+SR was validated on 600 low-resolution volumes, simulated from the corresponding high-resolution volumes. }
\vspace{-5pt}
\resizebox{\columnwidth}{!}
{\begin{tabular}{ccccccccc}
\toprule
& \multicolumn{3}{c}{Dice Index ($\%$)}     &     & \multicolumn{3}{c}{Hausdorff Dist. (mm)}            \\
\cmidrule{2-4} \cmidrule{6-8} 
&   SSLLN-HR & SSLLN-LR+SR               & $p$-value & & SSLLN-HR         &    SSLLN-LR+SR        & $p$-value\\ 
\midrule
LVC & \textbf{0.960$\pm$0.015} & 0.940$\pm$0.024     & $p\ll$0.001&     	& \textbf{3.396$\pm$0.505}	& 4.045$\pm$0.675    & $p\ll$0.001\\ 
LVW & \textbf{0.879$\pm$0.030} & 0.863$\pm$0.049     & $p\ll$0.001&     	& \textbf{3.868$\pm$1.306} 	& 4.394$\pm$0.841    & $p\ll$0.001\\ 
RVC & \textbf{0.929$\pm$0.025} & 0.914$\pm$0.033     & $p\ll$0.001&      	& \textbf{4.560$\pm$1.040}  & 5.039$\pm$1.218    & $p\ll$0.001\\
RVW & \textbf{0.662$\pm$0.103} & 0.557$\pm$0.121     & $p\ll$0.001&   	    & \textbf{5.664$\pm$2.701}	& 6.119$\pm$2.956 	 & $p\ll$0.001\\ 
\bottomrule
\end{tabular}}
\label{tb:2D3D}
\end{table}

In Table~\ref{tb:measurments}, we report the mean and standard deviation of the measurements derived from the two automated methods and manual segmentation. The table further demonstrates SSLLN-LR+SR results are comparable to SSLLN-HR results, proving that our proposed method can produce results comparable to direct segmentation of artefact-free HR volumes, even though target segmentation volumes are of low resolution and contain artefacts. Moreover, the RVM measurement derived from the two methods is consistent with the manual RVM measurement, confirming adequate segmentation of RVW using the two methods despite relatively lower Dice scores, as shown in Table~\ref{tb:2D3D}

\begin{table}[h!]
\vspace{-10pt}
\centering
\caption{Comparison of clinical measures between SSLLN-HR, SSLLN-LR+SR and manual measurements on 600 volumetric cardiac images. SSLLN-HR was validated on high-resolution volumes from Dataset 1, whilst SSLLN-LR+SR was validated on 600 low-resolution volumes, simulated from the corresponding high-resolution volumes. The 4th/5th columns show absolute difference between automated and manual measures.}
\vspace{-5pt}
\resizebox{\columnwidth}{!}
{\begin{tabular}{lccccc}
\toprule
&SSLLN-HR$^*$             & SSLLN-LR+SR$^\dagger$           & Manual            & $*$ vs Manual  & $\dagger$ vs Manual           	  \\ 
\midrule
LVV (ml) 	&{148.392$\pm$34.352} & 151.048$\pm$35.016    & 147.638$\pm$34.711   & 4.623$\pm$4.014  & 6.066$\pm$5.782  \\ 
LVM (gram)  &{123.028$\pm$24.123} & 124.240$\pm$24.383    & 119.278$\pm$25.685   & 5.551$\pm$4.308  & 6.737$\pm$5.244 \\ 
RVV (ml)  	&{168.638$\pm$37.144} & 174.383$\pm$39.480    & 171.553$\pm$38.622   & 8.547$\pm$7.540  & 9.299$\pm$8.551 \\
RVM (gram)  &{35.466$\pm$8.121}   & 32.290$\pm$7.381      & 33.704$\pm$7.261     & 3.571$\pm$2.803  & 2.956$\pm$2.899 \\ 
\bottomrule
\end{tabular}}
\label{tb:measurments}
\end{table}

\begin{table}[h!]
\vspace{-10pt}
\centering
\caption{Comparison of Dice index and Hausdorff distance between the proposed SSLLN-LR+SR and 5 state-of-the-art 3D approaches. These methods were tested on 20 simulated LR volumes ($\sim$200 CMR images). The ground-truth labels were obtained from high-resolution volumes acquired from same subjects, which do not contain cardiac artefacts.}
\vspace{-5pt}
\resizebox{\columnwidth}{!}
{\begin{tabular}{lcccccc}
\toprule
& \multicolumn{2}{c}{Endocardium}     &     & \multicolumn{2}{c}{Myocardium}            \\
\cmidrule{2-3} 						                    	\cmidrule{5-6} 
& Dice Index ($\%$) 	  				 & Hausdorff Dist. (mm)&          & Dice Index ($\%$)         & Hausdorff Dist. (mm)                  \\ 
\midrule
3D-Seg \cite{oktay2018anatomically} 	& {0.923$\pm$0.019}   & 10.28$\pm$8.25      &     		& {0.773$\pm$0.038}	    & 10.15$\pm$10.58    \\ 
3D-UNet \cite{cciccek20163d} 			& {0.923$\pm$0.019}   & 9.94$\pm$9.92       &       	& {0.764$\pm$0.045} 	& 9.81$\pm$11.77   \\ 
3D-AE \cite{ravishankar2017learning}	& {0.926$\pm$0.019}   & 8.42$\pm$3.64       &      		& {0.779$\pm$0.033}     & 8.52$\pm$2.72   \\
3D-ACNN \cite{oktay2018anatomically} 	& {{0.939$\pm$0.017}}   & 7.89$\pm$3.83       &   	    & {0.811$\pm$0.027}	    & 7.31$\pm$3.59	 \\ 
MAM \cite{bai2013probabilistic}			& {{0.87$\pm$0.029}}   & 6.65$\pm$1.74       &   	    & {0.711$\pm$0.064}	    & 8.89$\pm$2.07	 \\ 
SSLLN-LR+SR 		      & \textbf{{0.943$\pm$0.020}}   & \textbf{4.09$\pm$0.69}  &   & \textbf{{0.854$\pm$0.042}}     & \textbf{4.37$\pm$1.04}	 \\ 
\bottomrule
\end{tabular}}
\label{tb:3Dcomparison}
\end{table}

Next, we compare SSLLN-LR+SR with the 3D-seg model \cite{oktay2018anatomically}, 3D-UNet model \cite{cciccek20163d}, cascaded 3D-UNet and convolutional auto-encoder model (3D-AE) \cite{ravishankar2017learning}, 3D anatomically constrained neural network model (3D-ACNN) \cite{oktay2018anatomically} as well as multi-atlas method\footnote{Code is publicly available at \href{https://github.com/baiwenjia/CIMAS}{https://github.com/baiwenjia/CIMAS}} (MAM) \cite{bai2013probabilistic}. To ensure a fair comparison, we used the same 20 CMR volumes as in \cite{oktay2018anatomically} and the quantitative results are summarised in Table~\ref{tb:3Dcomparison}. Since 3D-ACNN only segments the left ventricle (LV), the table only shows the results for the endocardium and myocardium of LV. Among the methods compared, 3D-seg and 3D-UNet do not use shape information, while 3D-AE and 3D-ACNN infer shape constraints using an auto-encoder during network training. As Dice shows, MAM is inferior to deep learning-based methods, shape-based models outperform those without shape priors, and our SSLLN-LR+SR achieved the best performance. We propose three main reasons for this: 1): SSLLN-LR+SR uses atlas propagation to impose a shape refinement explicitly while 3D-AE and 3D-ACNN impose shape constraints in an implicit fashion. When the initial segmentation by SSLLN-LR is of sufficiently adequate quality, such an explicit shape refinement is able to produce more accurate segmentation. 2): SSLLN-LR+SR is a 2.5D-based method which allows the use of deeper network architectures than the 3D-based methods (e.g. ACNN-seg only uses 7 convolutional layers while SSLLN-LR+SR has 15), leading to improved segmentation accuracy. 3): SSLLN-LR+SR uses label-based non-rigid registration (\ref{eq:LC}), which may be more accurate for segmentation purpose than the intensity-based non-rigid registration used in MAM.

\subsection{Experiments on pathological low-resolution volumes}
\label{pathological}
\begin{figure}[h!]
\centering
\includegraphics[width=0.48\textwidth]{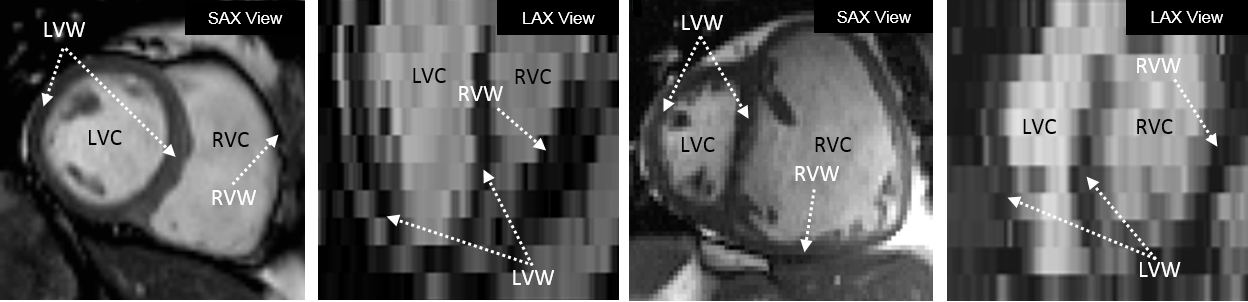}\\
\vspace{-5pt}
\caption{Illustrating the difference between a healthy subject (first two) and a PH subject (last two) from short- and long-axis views. Both subjects were scanned using low-resolution acquisition.}
\label{fig:PHhealth}
\end{figure}

In Section \ref{simluated}, we have quantitatively studied the performance of the proposed SSLLN-LR+SR using simulated LR cardiac volumes. In this section, we will use real LR volumes. In particular, we test SSLLN-LR+SR on volumetric data in patients with pulmonary hypertension (PH) in Dataset 2. PH leads to a progressive deterioration in cardiac function and ultimately death, due to RV failure. As such, it is critical to accurately segment different functional regions of the heart in PH so as to study PH patients quantitatively. Fig~\ref{fig:PHhealth} shows the difference in two CMR volumes from a representative healthy subject and a PH subject. In health, the RV is crescentic in short-axis views and triangular in long-axis views, wrapping around the thicker-walled LV. In PH, the dilated RV pushes onto the LV causing deformation and loss of its circular shape. The abnormal cardiac morphology of PH heart poses challenges for existing segmentation algorithms.

For training and testing, we use Dataset 2 introduced in Section \ref{clinicaldataset}. This dataset includes 629 LR PH volumes and 20 pairs of LR and HR PH volumes. We randomly split the 629 volumes into two disjoint subset of 429/200. The first subset is used to train SSLLN-LR, while the second subset is used for visually testing the accuracy of SSLLN-LR+SR segmentations (due to lack of corresponding HR ground truths). The 20 LR volumes are also used to quantitatively evaluate SSLLN-LR+SR using their HR volumes as ground-truth references. 231 HR atlases appearing in Section~\ref{HRExp} are used to refine SSLLN-LR segmentations. 

{200 greyscale PH volumes ($1.38 \times 1.38 \times 10$ mm) were segmented by SSLLN-LR+SR into HR smooth models ($1.25 \times 1.25 \times 2$ mm). Results were visually assessed by one clinician with over five years' experience in CMR imaging and judged satisfactory in all cases. We propose three reasons why the shape refinement works for PH cases: 1) the landmark-based affine and non-rigid registrations are collectively able to capture both global and local deformations between subjects; 2) for the non-rigid registration, we used label consistency as a loss function (\ref{eq:LC}). It is based on segmentation masks, which can provide stronger regional and edge information for an accurate registration; 3) multiple atlases (i.e. the most similar to the subject) were selected for registration and fusion, and these selected atlases together vote for the final result, which further prevents diseased cases producing healthy results.}

\begin{figure}[h!]
\vspace{-5pt}
\centering  
{\includegraphics[width=0.48\textwidth]{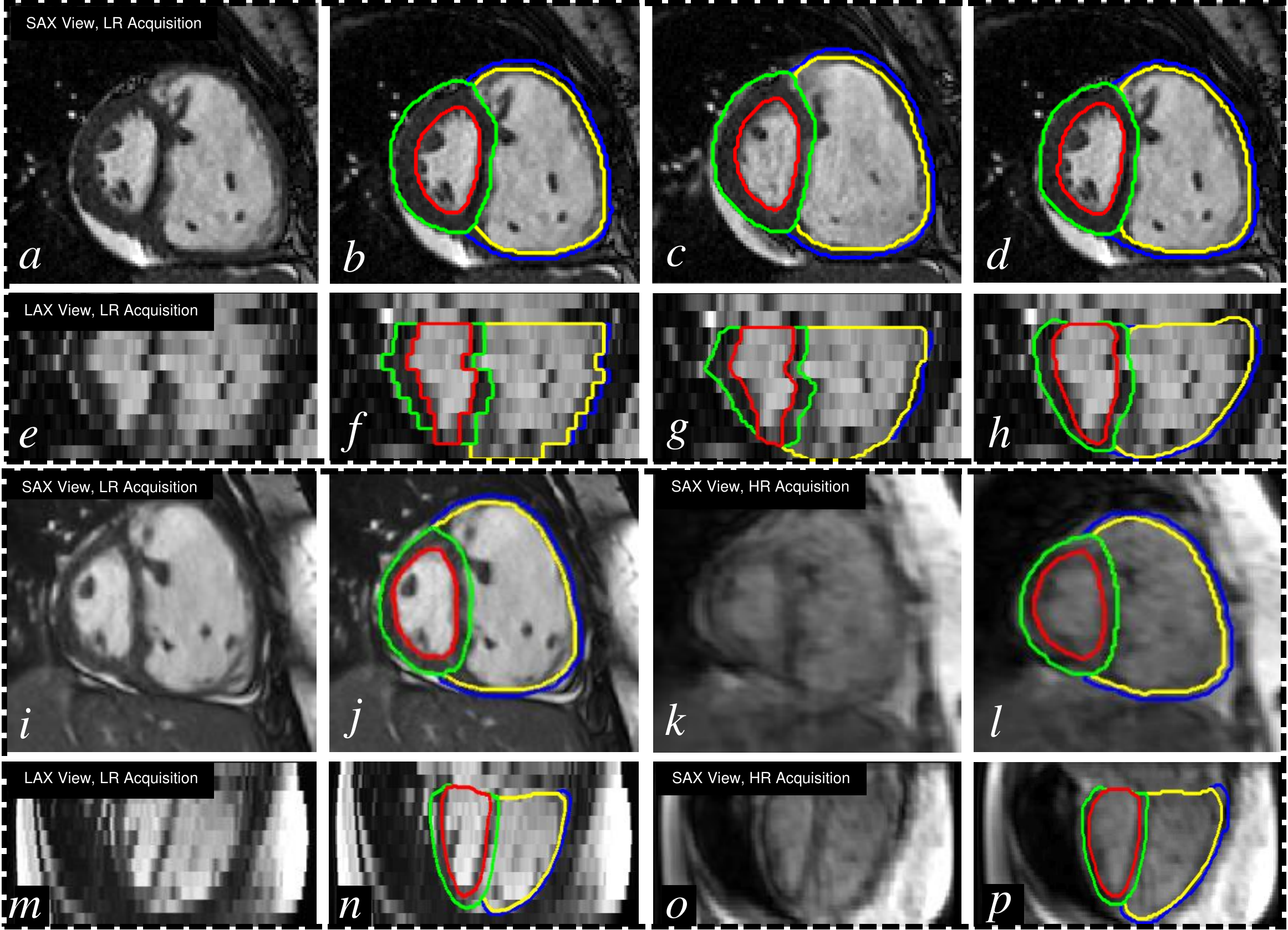}}\\%
\vspace{-5pt}%
\caption{Bi-ventricular segmentation of volumetric images from two PH patients. $a$ and $e$: original low-resolution volume (two views) from patient I; $b$ and $f$: 2D FCN+NNI results; $c$ and $g$: 2D FCN+SBI results; $d$ and $h$: SSLLR-LR+SC results. $i$ and $m$: original low-resolution volume from patient II; $j$ and $n$: SSLLN-LR+SR results; $k$ and $o$: original high-resolution volume from patient II; $l$ and $p$: ground truth. The proposed SSLLN-LR+SR is not only insensitive to cardiac artefacts (inter-slice shift, large slice thickness, and lack of slice coverage), but also robust against pathology-induced morphological changes.}
\label{fig:PHcurves}
\end{figure}

\begin{figure}[h!]
\vspace{-5pt}
\centering  
{\includegraphics[width=0.48\textwidth]{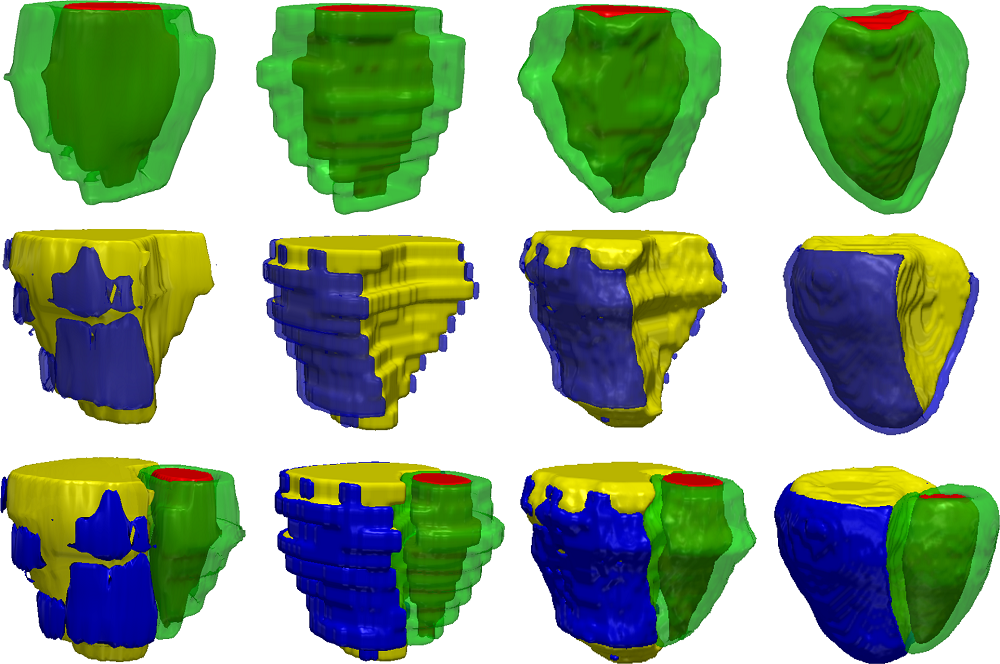}}\\
\caption{Visualisation of a 3D bi-ventricular model obtained through segmenting the volumetric image from a PH patient. 1st column: 2D FCN results; 2nd column: 2D FCN+NNI results; 3rd column: 2D FCN+SBI results; 4th colunm: SSLLR-LR+SC results. The proposed approach is capable of producing accurate, high-resolution and anatomically smooth bi-ventricular models for pathological subjects.}
\label{fig:visual3D}
\end{figure}

In Fig~\ref{fig:PHcurves} $a$-$h$ and Fig~\ref{fig:visual3D}, we show an exemplary bi-ventricular segmentation of a cardiac volume in PH. We visually compare SSLLN-LR+SR with 2D FCN \cite{bai2017human} and two approaches (nearest neighbour interpolation (NNI) and shape-based interpolation (SBI) \cite{frangi2002automatic,raya1990shape}) that interpolate the 2D FCN results. Both 2D FCN and interpolation methods do not use anatomical shape information, so they performed worse than SSLLN-LR+SR in the long-axis view, as confirmed in Fig~\ref{fig:PHcurves} $f$-$h$. Due to the high in-plane resolution, similar results in the short-axis view were achieved by different methods, as shown in Fig~\ref{fig:PHcurves} $b$-$d$. Moreover, we observed from Fig~\ref{fig:visual3D} that SSLLN-LR+SR gives a better 3D phenotype result which is smooth, accurate and artefact-free.

Next, we test SSLLN-LR+SR using 20 pairs of LR and HR cardiac volumetric images. In Fig~\ref{fig:PHcurves} $i$-$p$, we first demonstrate a segmentation example on a pair of LR and HR volumes acquired from the same patient with PH. The original low-resolution volume ($1.38 \times 1.38 \times 10$ mm) was segmented by SSLLN-LR+SR into a HR smooth model ($1.25 \times 1.25 \times 2$ mm). The smooth segmentation is then visually compared with the ground truth, obtained directly from segmenting the corresponding HR volume of the patient. As is evident, the paired segmentation results show a very good agreement in terms of their cardiac morphology. Further, Table~\ref{tb:PHmeasurments} is provided, which shows a quantitative comparison between the SSLLN-LR+SR results and the ground-truth segmentations. The automated measurements are quantitatively consistent with the manual measurements. Comparing Table~\ref{tb:PHmeasurments} with Table~\ref{tb:measurments}, we observed that PH patients have a bigger RVC and a smaller LVC than healthy subjects, and that the RVW of PH patients is thicker than that of healthy subjects. {Note that the Dice scores computed from the paired LR and HR volumes are not applicable here due to the fact that they were acquired from subjects scanned at different positions with different breath-holds.} We also note that $p$ values in Table~\ref{tb:PHmeasurments} are relatively large. This is likely due to the relatively low sample size of the dataset used in this experiment, in addition to the fact that automatic and manual measurements are not substantially different.

\begin{table}[h!]
\vspace{-10pt}
\centering
\caption{Comparison of clinical measures derived from SSLLN-LR+SR and manual segmentations on 20 pairs of low-resolution and high-resolution volumetric images from Dataset 2. SSLLN-LR+SR segmented 20 low-resolution volumes into high-resolution models, whilst manual segmentation was performed on 20 high-resolution cardiac volumes directly. The 4th column shows absolute difference between SSLLN-LR+SR and manual measures.}
\vspace{-5pt}
\resizebox{\columnwidth}{!}
{\begin{tabular}{lcccc}
\toprule
             & SSLLN-LR+SR$^\dagger$           & Manual             & $\dagger$ vs Manual   &$p$-value \\ 
\midrule
LVV (ml) 	 & 120.098$\pm$20.822 & 114.815$\pm$25.099    & 10.42$\pm$9.378     	 &$p\approx$0.119\\ 
LVM (gram)   & 125.989$\pm$34.639 & 124.237$\pm$25.271    & 8.032$\pm$9.614    	 &$p\approx$0.855\\ 
RVV (ml)  	 & 221.514$\pm$64.534 & 204.293$\pm$58.534    & 18.69$\pm$13.55   	 &$p\approx$0.001\\
RVM (gram)   & 51.621$\pm$14.938  & 49.877$\pm$14.166     & 3.857$\pm$2.558       &$p\approx$0.501\\ 
\bottomrule
\end{tabular}}
\label{tb:PHmeasurments}
\end{table}


\section{Discussion and Conclusion}
\label{conclusion}
In this paper, we developed a fully automatic pipeline for shape-refined bi-ventricular segmentation of short-axis CMR volumes. In the pipeline, we proposed a network that learns segmentation and landmark localisation tasks simultaneously. The proposed network combines the computational advantage of 2D networks and the capability of addressing 3D spatial consistency issues without loss of segmentation accuracy. The pipeline also induces an explicit shape prior information, thus allowing accurate, smooth and anatomically meaningful bi-ventricular segmentations despite artefacts in the cardiac volumes. Extensive experiments were conducted to validate the effectiveness of the proposed pipeline for both healthy and pathological cases.

{However, there still exist limitations in the pipeline. For example, the pipeline is a 2-stage approach, which is not end-to-end learning. In such a case, the network parameters learned in stage 1 might not be optimal to generate high-resolution smooth segmentations in stage 2. In addition, although the deployment of a trained network (SSLLN-HR or SSLLN-LR) in stage 1 took less than 1s, the shape refinement (SR) in stage 2 is relatively computationally expensive, which is a big disadvantage. SR combines the computational costs from atlas selection, target-to-atlas non-rigid image registration, and non-local label fusion. In our implementation, SR was performed in parallel for  5 selected atlases using multiple CPUs of a workstation and it took 15-20 mins per subject at ED. }


{In future work, we will investigate how to train a single network to compute smooth shapes from artefact-corrupted low-resolution cardiac volumes. A simple solution would be training an end-to-end super-resolution network, as in \cite{oktay2016multi}, but with the segmentation labels acquired from our pipeline as the ground truth inputs. We will also investigate how to improve the computational speed of Stage 2 in our pipeline. For example, a GPU-based non-rigid image registration toolbox\footnote{\href{http://cmictig.cs.ucl.ac.uk/research/software/software-nifty/niftyreg}{http://cmictig.cs.ucl.ac.uk/research/software/software-nifty/niftyreg}} could be utilised. Besides the GPU-based implementation, deep hashing \cite{erin2015deep,yun2018densely} may be explored to select relevant atlas subjects instead of the brute force search technique (i.e. nearest neighbour) currently used in our atlas selection process. Another direction will be to investigate how to adapt the proposed network architecture for different tasks. For example, a fully connected layer may be concatenated for classification of subjects into healthy versus pathological groups, which will be carried out simultaneously with segmentation and landmark localisation tasks. Our pipeline treats landmarks as voxels and classifies them. In future work, we will explore an alternative approach that treats landmarks as points and regresses their coordinates, which could be implemented with a fully connected layer.  }

%
%
%

%
%
%
%
%
%
%
%
%
\section{Acknowledgements}
The research was supported by the British Heart Foundation (NH/17/1/32725, RE/13/4/30184); the
EPSRC SmartHeart Programme (EP/P001009/1); the National Institute for Health Research (NIHR) Biomedical Research Centre based at Imperial College Healthcare NHS Trust and Imperial College London; and the Medical Research Council, UK. We would like to thank Dr Simon Gibbs, Dr Luke Howard and Prof Martin Wilkins for providing the CMR image data. The TITAN Xp GPU used for this research was kindly donated by the NVIDIA Corporation.

\bibliographystyle{ieeetr}
\bibliography{ref}

\begin{thebibliography}{10}

\bibitem{ripley2016cardiovascular}
D.~Ripley, T.~Musa, L.~Dobson, S.~Plein, and J.~Greenwood, ``Cardiovascular
  magn. reson. imaging: what the general cardiologist should know,'' {\em
  Heart}, pp.~heartjnl--2015, 2016.

\bibitem{medrano2015challenges}
P.~Gracia, B.~Cowan, A.~Suinesiaputra, and A.~Young, ``Challenges of cardiac
  image analysis in large-scale population-based studies,'' {\em Curr. Cardiol.
  Rev.}, vol.~17, no.~3, p.~9, 2015.

\bibitem{rueckert2016learning}
D.~Rueckert, B.~Glocker, and B.~Kainz, ``Learning clinically useful information
  from images: Past, present and future,'' {\em Med. Image Anal.}, vol.~33,
  pp.~13--18, 2016.

\bibitem{winther2017nu}
H.~Winther, C.~Hundt, B.~Schmidt, C.~Czerner, J.~Bauersachs, F.~Wacker, and
  J.~Vogel, ``V-net: Deep learning for generalized biventricular cardiac mass
  and function parameters,'' {\em ArXiv Preprint ArXiv:1706.04397}, 2017.

\bibitem{patravali20172d}
J.~Patravali, S.~Jain, and S.~Chilamkurthy, ``2d-3d fully convolutional neural
  networks for cardiac mr segmentation,'' {\em ArXiv Preprint
  ArXiv:1707.09813}, 2017.

\bibitem{baumgartner2017exploration}
C.~Baumgartner, L.~Koch, M.~Pollefeys, and E.~Konukoglu, ``An exploration of 2d
  and 3d deep learning techniques for cardiac mr image segmentation,'' {\em
  ArXiv Preprint ArXiv:1709.04496}, 2017.

\bibitem{isensee2017automatic}
F.~Isensee, P.~Jaeger, P.~Full, I.~Wolf, S.~Engelhardt, and K.~Maier,
  ``Automatic cardiac disease assessment on cine-mri via time-series
  segmentation and domain specific features,'' {\em ArXiv Preprint
  ArXiv:1707.00587}, 2017.

\bibitem{zheng20183d}
Q.~Zheng, N.~Delingette, Hand~Duchateau, and N.~Ayache, ``3d consistent \&
  robust segmentation of cardiac images by deep learning with spatial
  propagation,'' {\em IEEE T. Med. Imaging}, 2018.

\bibitem{nasr2018left}
M.~Nasr, M.~Mohrekesh, M.~Akbari, S.~Soroushmehr, E.~Nasr, N.~Karimi,
  S.~Samavi, and K.~Najarian, ``Left ventricle segmentation in cardiac mr
  images using fully convolutional network,'' {\em ArXiv Preprint
  ArXiv:1802.07778}, 2018.

\bibitem{khened2018fully}
M.~Khened, V.~Kollerathu, and G.~Krishnamurthi, ``Fully convolutional
  multi-scale residual densenets for cardiac segmentation and automated cardiac
  diagnosis using ensemble of classifiers,'' {\em ArXiv Preprint
  ArXiv:1801.05173}, 2018.

\bibitem{oktay2018anatomically}
O.~Oktay, E.~Ferrante, K.~Kamnitsas, M.~Heinrich, W.~Bai, J.~Caballero,
  S.~Cook, A.~Marvao, T.~Dawes, D.~O‘Regan, {\em et~al.}, ``Anatomically
  constrained neural networks (acnns): application to cardiac image enhancement
  and segmentation,'' {\em IEEE T. Med. Imaging}, vol.~37, no.~2, pp.~384--395,
  2018.

\bibitem{tran2016fully}
P.~Tran, ``A fully convolutional neural network for cardiac segmentation in
  short-axis mri,'' {\em ArXiv Preprint ArXiv:1604.00494}, 2016.

\bibitem{bai2017human}
W.~Bai, M.~Sinclair, G.~Tarroni, O.~Oktay, M.~Rajchl, G.~Vaillant, A.~Lee,
  N.~Aung, E.~Lukaschuk, M.~Sanghvi, {\em et~al.}, ``Human-level cmr image
  analysis with deep fully convolutional networks,'' {\em J. Cardiov. Magn.
  Reson.}, 2018.

\bibitem{ngo2017combining}
T.~Ngo, Z.~Lu, and G.~Carneiro, ``Combining deep learning and level set for the
  automated segmentation of the left ventricle of the heart from cardiac cine
  magnetic resonance,'' {\em Med. Image Anal.}, vol.~35, pp.~159--171, 2017.

\bibitem{avendi2016combined}
M.~Avendi, A.~Kheradvar, and H.~Jafarkhani, ``A combined deep-learning and
  deformable-model approach to fully automatic segmentation of the left
  ventricle in cardiac mri,'' {\em Med. Image Anal.}, vol.~30, pp.~108--119,
  2016.

\bibitem{duan2018deep}
J.~Duan, J.~Schlemper, W.~Bai, T.~J. Dawes, G.~Bello, G.~Doumou, A.~De~Marvao,
  D.~P. O’Regan, and D.~Rueckert, ``Deep nested level sets: Fully automated
  segmentation of cardiac mr images in patients with pulmonary hypertension,''
  in {\em MICCAI}, pp.~595--603, Springer, 2018.

\bibitem{schlemper2018cardiac}
J.~Schlemper, O.~Oktay, W.~Bai, D.~C. Castro, J.~Duan, C.~Qin, J.~V. Hajnal,
  and D.~Rueckert, ``Cardiac mr segmentation from undersampled k-space using
  deep latent representation learning,'' in {\em MICCAI}, pp.~259--267,
  Springer, 2018.

\bibitem{bernard2018deep}
O.~Bernard, A.~Lalande, C.~Zotti, {\em et~al.}, ``Deep learning techniques for
  automatic mri cardiac multi-structures segmentation and diagnosis: Is the
  problem solved?,'' {\em IEEE T. Med. Imaging}, 2018.

\bibitem{petersen2015uk}
S.~Petersen, P.~Matthews, J.~Francis, M.~Robson, F.~Zemrak, R.~Boubertakh,
  A.~Young, S.~Hudson, P.~Weale, S.~Garratt, {\em et~al.}, ``Uk biobank’s
  cardiovascular magnetic resonance protocol,'' {\em J. Cardiov. Magn. Reson.},
  vol.~18, no.~1, p.~8, 2015.

\bibitem{tarroni2018learning}
G.~Tarroni, O.~Oktay, W.~Bai, A.~Schuh, {\em et~al.}, ``Learning-based quality
  control for cardiac mr images,'' {\em arXiv preprint arXiv:1803.09354}, 2018.

\bibitem{oktay2016multi}
O.~Oktay, W.~Bai, M.~Lee, {\em et~al.}, ``Multi-input cardiac image
  super-resolution using convolutional neural networks,'' in {\em MICCAI},
  pp.~246--254, Springer, 2016.

\bibitem{petitjean2011review}
C.~Petitjean and J.~Dacher, ``A review of segmentation methods in short axis
  cardiac mr images,'' {\em Med. Image Anal.}, vol.~15, no.~2, pp.~169--184,
  2011.

\bibitem{de2014population}
A.~Marvao, T.~Dawes, W.~Shi, C.~Minas, N.~Keenan, T.~Diamond, G.~Durighel,
  G.~Montana, D.~Rueckert, and S.~o. Cook, ``Population-based studies of
  myocardial hypertrophy: high resolution cardiovascular magnetic resonance
  atlases improve statistical power,'' {\em J. Cardiov. Magn. Reson.}, vol.~16,
  no.~1, p.~16, 2014.

\bibitem{milletari2016v}
F.~Milletari, N.~Navab, and S.~Ahmadi, ``V-net: Fully convolutional neural
  networks for volumetric medical image segmentation,'' in {\em 2016 Fourth
  International Conference on 3D Vision}, pp.~565--571, IEEE, 2016.

\bibitem{kamnitsas2017efficient}
K.~Kamnitsas, C.~Ledig, V.~F. Newcombe, {\em et~al.}, ``Efficient multi-scale
  3d cnn with fully connected crf for accurate brain lesion segmentation,''
  {\em Med. Image Anal.}, vol.~36, pp.~61--78, 2017.

\bibitem{mortazi2017cardiacnet}
A.~Mortazi, R.~Karim, K.~Rhode, J.~Burt, and U.~Bagci, ``Cardiacnet:
  Segmentation of left atrium and proximal pulmonary veins from mri using
  multi-view cnn,'' in {\em MICCAI}, pp.~377--385, Springer, 2017.

\bibitem{grbic2012complete}
S.~Grbic, R.~Ionasec, D.~Vitanovski, I.~Voigt, Y.~Wang, B.~Georgescu, N.~Navab,
  and D.~Comaniciu, ``Complete valvular heart apparatus model from 4d cardiac
  ct,'' {\em Med. Image Anal.}, vol.~16, no.~5, pp.~1003--1014, 2012.

\bibitem{duan2018combining}
J.~Duan, J.~Schlemper, W.~Bai, T.~J. Dawes, G.~Bello, C.~Biffi, G.~Doumou,
  A.~De~Marvao, D.~P. O’Regan, and D.~Rueckert, ``Combining deep learning and
  shape priors for bi-ventricular segmentation of volumetric cardiac magnetic
  resonance images,'' in {\em International Workshop on Shape in Medical
  Imaging}, pp.~258--267, Springer, 2018.

\bibitem{rueckert1999nonrigid}
D.~Rueckert, L.~Sonoda, C.~Hayes, D.~Hill, M.~Leach, and D.~Hawkes, ``Nonrigid
  registration using free-form deformations: application to breast mr images,''
  {\em IEEE T. Med. Imaging}, vol.~18, no.~8, pp.~712--721, 1999.

\bibitem{frangi2002automatic}
A.~Frangi, D.~Rueckert, J.~Schnabel, and W.~Niessen, ``Automatic construction
  of multiple-object three-dimensional statistical shape models: Application to
  cardiac modeling,'' {\em IEEE T. Med. Imaging}, vol.~21, no.~9,
  pp.~1151--1166, 2002.

\bibitem{buades2005non}
A.~Buades, B.~Coll, and J.~Morel, ``A non-local algorithm for image
  denoising,'' in {\em CVPR}, vol.~2, pp.~60--65, IEEE, 2005.

\bibitem{duan2015fast}
J.~Duan, Z.~Pan, B.~Zhang, W.~Liu, and X.-C. Tai, ``Fast algorithm for color
  texture image inpainting using the non-local ctv model,'' {\em J. Global
  Optim.}, vol.~62, no.~4, pp.~853--876, 2015.

\bibitem{lu2019graph}
W.~Lu, J.~Duan, D.~Orive-Miguel, L.~Herve, and I.~B. Styles, ``Graph-and finite
  element-based total variation models for the inverse problem in diffuse
  optical tomography,'' {\em Biomed. Opt. Express}, vol.~10, no.~6,
  pp.~2684--2707, 2019.

\bibitem{bai2015multi}
W.~Bai, W.~Shi, C.~Ledig, and D.~Rueckert, ``Multi-atlas segmentation with
  augmented features for cardiac mr images,'' {\em Med. Image Anal.}, vol.~19,
  no.~1, pp.~98--109, 2015.

\bibitem{cciccek20163d}
{\"O}.~{\c{C}}i{\c{c}}ek, A.~Abdulkadir, S.~Lienkamp, T.~Brox, and
  O.~Ronneberger, ``3d u-net: Learning dense volumetric segmentation from
  sparse annotation,'' in {\em MICCAI}, pp.~424--432, Springer, 2016.

\bibitem{ravishankar2017learning}
H.~Ravishankar, R.~Venkataramani, S.~Thiruvenkadam, P.~Sudhakar, and V.~Vaidya,
  ``Learning and incorporating shape models for semantic segmentation,'' in
  {\em MICCAI}, pp.~203--211, Springer, 2017.

\bibitem{bai2013probabilistic}
W.~Bai, W.~Shi, and othersl, ``A probabilistic patch-based label fusion model
  for multi-atlas segmentation with registration refinement: application to
  cardiac mr images,'' {\em IEEE T. Med. Imaging}, vol.~32, no.~7,
  pp.~1302--1315, 2013.

\bibitem{raya1990shape}
S.~Raya and J.~Udupa, ``Shape-based interpolation of multidimensional
  objects,'' {\em IEEE T. Med. Imaging}, vol.~9, no.~1, pp.~32--42, 1990.

\bibitem{erin2015deep}
V.~Erin, J.~Lu, G.~Wang, P.~Moulin, and J.~Zhou, ``Deep hashing for compact
  binary codes learning,'' in {\em CVPR}, pp.~2475--2483, 2015.

\bibitem{yun2018densely}
Y.~Gu and J.~Yang, ``Densely-connected multi-magnification hashing for
  histopathological image retrieval,'' {\em IEEE J. Biomed. Health Inform.},
  2018.

\end{thebibliography}

\end{document}